\documentclass[journal]{IEEEtran}
\usepackage{amsmath,amsfonts}
\usepackage{mathtools}
\usepackage{algorithm}
\usepackage{algpseudocode}
\usepackage{amssymb}
\usepackage{setspace}
\usepackage{array}
\usepackage[font=normalsize]{subfig}
\usepackage{textcomp}
\usepackage{stfloats}
\usepackage{url}
\usepackage{verbatim}
\usepackage{graphicx}
\usepackage{tikz}
\usetikzlibrary{arrows.meta,decorations.markings,calc,patterns}
\usepackage{cite}
\usepackage{pgf} 
\usepackage{doi}
\DeclarePairedDelimiter\abs{\lvert}{\rvert}%
\DeclarePairedDelimiter\norm{\lVert}{\rVert}%
\makeatletter
\let\oldabs\abs
\def\abs{\@ifstar{\oldabs}{\oldabs*}}
\let\oldnorm\norm
\def\norm{\@ifstar{\oldnorm}{\oldnorm*}}
\makeatother
\renewcommand{\vec}[1]{\boldsymbol{#1}}

\DeclareMathOperator*{\argmin}{argmin}

\DeclareMathOperator{\Exp}{Exp}
\DeclareMathOperator{\Log}{Log}
\usepackage{flushend}

\begin{document}

\title{Motion-Acceleration Calibration and Compensation in IMUs \\ without External Equipment for Attitude Estimation Filters}


\author{Fabian Arzberger and Andreas N{\"u}chter} 



\maketitle

\begin{abstract}
  Attitude estimation based on inertial sensing requires measurements of local angular velocities and local gravity via gyroscopes and accelerometers.
  However, during the motion of a mobile system the inertial measurement unit (IMU) will be subject to additional accelerations which skews the measurement of local gravity.
  This effect gets amplified the further away the IMU is from the base of the system.
  Many attitude estimation filters, such as ``Madgwick'' or ``Mahony'', account for this by relying more on gyroscope integration for periods of high angular velocity.
  However, this approach is prone to accumulate long term error especially around the gravity vector.
  In this work we utilize the gyroscope measurements to compensate the additional accelerations induced by the motion of the system, i.e., centripetal- and tangential-accelerations.
  Additionally, we introduce a calibration method that estimates intrinsic IMU parameters such as axes misalignment, bias, scale, as well as the extrinsic base-to-IMU vector without the necessity for additional external equipment.
  Our evaluation in simulation as well as in the real-world shows that this method improves any attitude filter that relies on the direction of gravity.
  Furthermore we demonstrate the effectivenes on highly dynamic systems, and systems that are unable to put the IMU at the center of rotation, using our real-world spherical mobile mapping system.
\end{abstract}


\section{Introduction}

Inertial measurement units (IMUs) are well-known and essential sensors for many types of mobile systems.
IMUs consist of accelerometer-, gyroscope-, and sometimes, magnetometer-triads.
They are vital instruments for attitude estimation and are used in countless robotic-, navigation-, and computer-vision-pipelines.
Some systems do not rely on the magnetometer, especially considering applications with strong external magnetic fields, or spaces, where a magnetic field is not always available.
The field of gravity, on the other hand, is a more reliably source of inertial direction measured by the accelerometer.
Since the rotation around the gravity direction is not observable by the accelerometers alone, one has to rely partially on either the magnetometers or measurements of the bodies angular velocity.
The gyroscopes measure anuglar velocity, which makes rotations around all 3 degrees of freedom (DoF) observable.
However, relying solely on integration of angular velocities leads to drift, which is why many approaches apply filtering algorithms to estimate the attitude from multiple modalities.    
During the motion of a mobile system, the rigidly mounted IMU will be subject not only to gravity, but to additional accelerations which skews the measurement.
This effect depends on the specific trajectory of the sensor, but generally gets amplified the further away the IMU is from the systems center of rotation.
Any rotation around that center leads to centripetal- and tangential-forces in the inertial frame, which the accelerometer then measures.
\begin{figure}
  \centering
  \begin{tikzpicture}
    \draw (-4,0) -- (4,0);
    \fill[pattern=north east lines] (-4,0) rectangle (4,-0.2);
    
    \coordinate (Origin) at (-3.9,0.1);
    \coordinate (Contact) at (0,0);
    \coordinate (Center) at (0,1.5);
    \coordinate (Sensor) at (-1.1,0.8);
    
    \draw (Center) circle (1.5cm);
    
    \fill[gray] (Sensor) circle (2pt) node[below left=2pt] {IMU};
    
    \fill (Center) circle;
    
    \draw[-{Stealth[length=2mm]},blue] (Origin) -- (Center) node[above] {$\vec{r}_c$};
    \draw[-{Stealth[length=2mm]},blue] (Center) -- (Sensor) node[below right] {$\vec{r}_e$};
    
    \draw[dashed] (Center) -- (-1.5, 1.5) node[midway,above] {$R$};
    
    \draw[-{Stealth[length=2mm]}] (3.5,0) -- (3.5,0.8) node[midway,left] {$\vec{n}$};
    
    \draw[-{Stealth[length=1.5mm]},thin] (-0.5,2.2) arc (170:10:0.5) node[pos=0.5, above=0.1pt] {$\vec{\omega}$};
    
    \draw[-{Stealth[length=2mm]},orange] (Sensor) -- ($(Sensor)!0.6!(Center)$) node[midway, right=4pt] {\small $\vec{\omega} \!\times\! (\vec{\omega} \!\times\! \vec{r}_e)$};
    
    \draw[-{Stealth[length=2mm]},green!60!black] (Sensor) -- ++(-0.35,0.5) node[left] {\small $\frac{\partial \vec{\omega}}{\partial t} \!\times\! \vec{r}_e$};
    
    \draw[-{Stealth[length=2.5mm]}] (1.3,0.1) -- (2.3,0.1) node[midway,above] {$R\vec{\omega} \!\times\! \vec{n}$};
    
    
    \draw[-{Stealth[length=1.5mm]}] (Origin) -- ++(0.6,0) node[right] {$x$};
    \draw[-{Stealth[length=1.5mm]}] (Origin) -- ++(0,0.6) node[above] {$z$};
    \fill[white] (Origin) circle (2pt);
    \draw (Origin) circle (2pt);
    \fill (Origin) circle (0.8pt);
    \node[below left=3pt] at (Origin) {$y$};
    
  \end{tikzpicture}
  \caption{Spherical system rolling on a surface with normal $\vec{n}$. 
  A rigidly mounted IMU (gray) is located at position $\vec{r}_e$ relative to the ball center $\vec{r}_c$. 
  The centripetal (orange) and tangential (green) accelerations act on the sensor due to the angular motion $\vec{\omega}$ around $\vec{r}_c$.}\label{fig:spherical_model}
\end{figure}
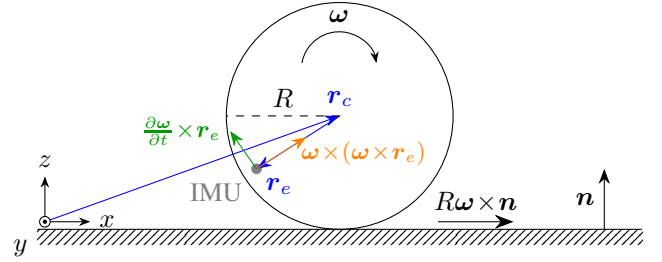
In this work, we put a particular emphasis to spherical systems as illustrated in Figure~\ref{fig:spherical_model}, which is one possible application for motion force compensation.
Many other examples exist where the same idea applies.
The effect of rotation related forces on IMUs have been identified early on, yet the center of rotation varies across systems.
For example, in 1969 the NASA considered the trajectory of a vertical and/or short take-off and landing (V/STOL) aircraft~\cite{ibm1969vstol_alt}.
Using the aircraft trajectory, which is estimated from the INS, RADAR, and control inputs, they calculate and subsequently subtract the centripetal- and coriolis-forces acting on the aircraft.
Their work omits the tangential-force, since the primary motion acceleration effect was due to a large distance to the center of rotation during turning flight, instead of fast rotary dynamics. 
Similarly, in a military technical report from 1979~\cite{hung1979size} the authors identify the effect of centripetal- and tangential-forces on the strapdown accelerometer in cruise missiles and label it the ``size effect''.  
They also identify a simple yet effective way to negate the effects due to the vehicles own body angular rate, which is to mount multiple IMUs symetrically around the missiles turning center and average their measurements. 
Another similar but more recent example is~\cite{euston2008complementary}, where the authors propose an attitude estimation filter for a fixed-wing unmanned aerial vehicle (UAV).
When the UAV banks in, the filter estimates the turn radius using bank angle and airspeed data, calculates the resulting centripetal acceleration and subtracts it from the accelerometer reading.   
While the idea of motion acceleration compensation exists primarily in the aerospace and military community, it is heavily underrepresented in the robotics community.
Since the centripetal- and tangential-accelerations depend on the angular velocity measurement and its derivative (c.f.,~Figure~\ref{fig:spherical_model}), one possible explanation is the comparably lower precision in micro-electro-mechanical systems (MEMS) gyroscopes.
Another explanation might be a lack of motivation for systems where the displacement to the center of rotation or magnitude of angular velocity are small. 
Whilst not applicable to all systems, there exist many examples where the non-centered placement of the IMU on the systems base is rigid, a center of rotation exists while turning, and motion acceleration effects are not negligable:
(1) Spherical mobile mapping, where the IMU is mounted rigidly inside the rolling ball, (2) robotic arms or grippers, where the IMU is mounted on the end-effector, (3) bio-inspired mobile robots, (4) rotary-wing UAVs, or (5) in-orbit spacecrafts and satellite systems.
Placing the IMU on the center of rotation might not be possible due to design constraints.
Generally, the measurement of local gravity gets more skewed the higher the angular velocity and the higher the displacement of the sensor towards the center of rotation is. 
Many attitude estimation filters account for this effect by putting more weight in the gyroscope integration for periods of higher angular velocity, leading to drift.
However, if the displacement is known, the attitude estimation is able to put more weight in the corrected accelerometer measurements to reduce drift.
In this work, we show that any attitude estimation filter benefits from doing this, and provide a robust IMU calibration method to recover the displacement.
The proposed method also calibrates the intrinsic scale, misalginment, and bias parameters of the triads and is able to recover the relative rotation between multiple IMUs.
No external equipment other than the mounting base of the system itself and no external reference other than the magnitude of local gravity is required.
We omit the usage of the magnetometer in this work due to the work on the
spherical mobile mapping system in extraterestrial exploration~\cite{deadalus}.
The contributions of this work are:
\begin{itemize}
  \item A robust method to calibrate intrinsic scale, misalginment, and bias parameters of IMU accelerometer and gyroscope triads, as well as the displacement vectors to the center of rotation and relative sensor orientations in a multi-IMU system, without external equipment.
  \item A numerically stable method for online motion acceleration compensation that subtracts centripetal- and tangential-terms in the accelerometer measurements using corresponding gyroscope measurements.
  \item An evaluation of the calibration and online compensation method using synthetic, semi-synthetic, and real-world datasets, showing that any state-of-the-art attitude estimation filter improves its accuracy by using the proposed method. 
  \item We open-source our calibration- and simulation-software, as well as calibration datasets from our spherical mobile mapping prototype.\footnote{\url{github.com/fallow24/ros_imu_calib}}
\end{itemize}

\section{Related Works}

The proposed calibration method is inspired by~\cite{imutk2014}, which is a robust method for intrinsic accelerometer and gyroscope triads calibration without external equippment. 
In the literature the data collection approach is commonly referred to as the ``multi-position method'',  consisting of two steps: 
First, the IMU system is static for a moderate duration.
The authors suggest $60$~s as a generally good value but note that sensor dependend values can be derived from plotting the Allan variance.
Then second, an operator has to manually put the IMU system in different orientations for a brief duration.
The authors state that $36$ to $50$ different attitudes for at least $5$~s are sufficient for good calibration. 
Our proposed calibration method does not change this data collection approach.
We extend their method, using the same data, to recover the displacement vector from the center of rotation towards the sensor, as well as the relative orientations between sensors in a multi-IMU system.
When collecting a dataset for calibration, the system should primarily experience rotations around one common center point.
However, when testing the proposed method with real-world handheld data, it proves to be robust against small deviations from this assumption.

\subsection{Attitude Estimation}
The de-facto standards of state-of-the-art attitude estimation filters are twofold:
(1) Many systems try placing the IMU as close to the center of rotation as possible to minimize centripetal and tangential contributions~\cite{6957468}, and 
(2) put less trust on the accerlerometer estimations during dynamic motions.
The latter happens if the local gravity vector predicted by gyroscope integration does not match the one measured by the accelerometer.
A popular family of ``non-optimal'' filters are extended Kalman filters (EKF), e.g., formulated with quaternions (QEKF)~\cite{qekf}.
In~\cite{6172226}, the authors propose a QEKF that also includes a dynamic acceleration term in the model equations.
EKF are generally considered non-optimal since the first-order linearization via Jacobians introduces truncation errors and forces a Gaussian approximation onto the inherently non-Gaussian posterior of the nonlinear system.
This breaks the mathematically proven minimum-variance optimality of the original linear Kalman filter (LKF)~\cite{kalmanoriginal}, but still works generally well in practice~\cite{crassidis2007survey}.
QEKF are especially prominent in unmanned aerlial vehicles (UAV)~\cite{jing2017attitude, zhang2017attitude, hall2008quaternion} and in-orbit spacecrafts~\cite{lefferts1982kalman}.
Two more recent examples include~\cite{wei2025robust} or~\cite{chen2024robust}, where the authors implement error-state Kalman filters (ESKF) for attitude estimation which include a general dynamic acceleration term in the error-state equations.
Unlike the EKF, an ESKF does not linearize the full state, but only the error-state.
Since the error state is close to zero this leads to better linearization accuracy and slower covariance divergence.
However, ESKF are still non-optimal. 
Iterated error-state Kalman filters (IESKF) improve on ESKF by re-linearizing the measurement model around the updated nominal state estimate over multiple iterations within each update step, effectively solving a Gauss-Newton problem at each timestep.
One example is~\cite{10203029}, where the authors propose an IESKF to estimate the linear non-gravity accelerations using the error-state equations.
Additionally, their approach includes a long short-term memory (LSTM) neural network which assists the IESKF by estimating the non-Gaussian process noise covariance. 
Another less popular extension of the Kalman filter for non-linear systems is the unscented Kalman filter (UKF)~\cite{julier1997new}.
Instead of linearizing the models with Jacobians, a discrete sampling approach using the unscented transform approximates the mean and covariance of the state distribution.
One example of a UKF implementation with quaternions for attitude estimation is~\cite{1257247}.
Generally, UKF achive higher-order accuracy and increased robustness compared to EKF, but are harder to tune effectively and are computationally more complex.
Many of the aforementioned works estimate dynamic accelerations indirectly, yet the centripetal- and tangential-accelerations are never explicitly modelled. 
Arguably the most commonly used filter due to its simplicity and good performance is the Complementary filter~\cite{complementary}.
It has usually only one dynamic parameter that weights the attitude estimations of the accerlerometer and gyroscope complementarily, hence the name.
The weight of the accelerometer estimation is then decreased heuristically by either considering the estimated gravity direction error or the magnitude of the angular velocity. 
Another common filter is the ``Mahony''-filter~\cite{mahony}.
This filter can be categorized as a non-linear explicit complementary filter that acts directly on the $SO(3)$ group and additionally estimates the gyroscope bias online.
A geometric observer estimates the gyroscope integration error on the manifold which is then used in a proportional-integral feedback scheme.
This is clearly motivated by control theory, i.e., the authors perform Lyapunov-stability analysis to ensure global stability in the observed error.  
The ``Madgwick''-filter~\cite{madgwick} also estimates the gyroscope integration error and corrects it with a gradient descent appraoch using accelerometer data.
The method needs to compute only four derivatives, one for each quaternion dimension, and is thus computationally efficient.
Furthermore, the Madgwick filter has only one parameter which is linked to the mean zero gyroscope error and is thus straightforward to tune.
Finally, the ``Autogain''-filter
is specialized for spherical systems.
It combines the well-known Complementary-~\cite{complementary} with the Madgwick-~\cite{madgwick} filter and automatically adjusts the filter gains based on the angular velocity. 

All of the abovementioned attitude estimation filters rely on the measured direction of gravity in the inertial frame, yet the IMU measurements also include acceleration due to motion. 
Consequently, a method that compensates for the additional accelerations during motion improves the estimation accuracy of any filter. 

\subsection{Motion Compensation}
Few approaches exist in the literature that are concerned with utilizing or even compensating the measured intertial acceleration due to motion.
In~\cite{guner2022novel} the authors propose a calibration method for non-centered IMUs based on rotations induced by an external servo.
The known angular velocities from the servo are used to predict the centripetal- and tangetial-accelerations which are then used to calibrate accelerometer intrinsic parameters. 
Furthermore, the method also estimates the position of the IMU with respect to the axis of rotation.
However, the authors focus only on rotations around one single axis at a time, and their method requires external equipment.
Another approach that considers the off-centered motion accelerations to obtain better gravity measurements is~\cite{9110127}.
Unlike our method,~\cite{9110127} considers the lever-arm as an intrinsic parameter of a highly precise fiber optical gyroscope (FOG) IMU.
Since the lever-arm is comparably small, the method requires mounting the IMU to an accurate turntable that captures data for around $30$~min. 
To the best of our knowledge, the most similar approach to ours is~\cite{9423615}:
The authors provide a calibration procedure for off-centered IMUs that estimates the intrinsic and extrinsic parameters by utilizing the motion accelerations. 
However, unlike our approach, their method optimizes all $45$ intrinsic and extrinsic paramters simultaneously in one optimization problem.
Additionally, there exist $4$ hyper-paramters which the authors estimate via the covariances of the underlying IMU data.
This makes their method sensitive to the initial guess: 
Selecting improper values leads to optimizing towards a local optimimum, which results in failed calibration.
Our approach does not suffer from this and is robust even from large initial deviations. 
Three additional differences to~\cite{9423615} are: 
(1) This work considers the intrinsic scale errors as linear instead of non-linear, 
(2) our data collection approach does not depend on free-fall data, i.e., does not require throwing the system in the air, and 
(3) the proposed method uses Derivative of Gaussian (DoG) kernels instead of 3rd-order finite-differences to calculate angular accelerations.
Finally, there is one remarkable example of motion acceleration compensation from the LiDAR-Inertial Odometry (LIO) community:
In~\cite{chen2023direct} the authors present ``Direct LiDAR-Inertial Odometry and Mapping'' (DLIOM) in an ArXiv preprint from 2023, which seems to be a continuation of the authors work in~\cite{chen2022direct}.
Notably, in the preprint DLIOM estimates the centripetal- and tangential-accelerations acting on the IMU due to the non-centered placement on the mounting base in Equation~4.
Strikingly however, in the corresponding published conference version~\cite{10160508} this Equation does not show up.
We do not certainly know why the authors have abandoned the idea.
A possible explanation is an issue with the angular acceleration needed for the tangential term, which has to be numerically derived using noisy gyroscope measurements. 
In the following sections of this work, we first derive the motion compensation model and show it is beneficial if the derivative of the angular velocity is handled carefully.
Based on this, we propose a calibration framework to estimate the displacement vector to the center of rotation, which is then used to correct the accelerometer measurment on-line. 

\section{Method}
It is our goal to correct the accelerometer measurement, i.e., subtract the accelerations due to linear and angular motion, such that only the gravity component remains. 
This benefits the downstream attitude estimation, which relies on the direction of gravity.
The presented procedure uses the same data recording approach as in~\cite{imutk2014}, i.e., we need to collect measurements where the sensor is static for a short duration ($\approx 5~s$) and then rotated into a different orientation.
This process is repeated around $36 - 50$ times.
Depending on the platform, a user moves the system in different orientations, or the system might do this itself. 
The calibration procedure then recovers the $SE(3)$ transformation between all IMUs.
Specifically, our method estimates the position vectors from the center of rotation towards the sensors, as well as the orientation between them.

\subsection{Derivation of the Model}\label{ssec:derivation}
In this section we derive the model describing the centripetal and tangential
forces acting on the sensor, resulting from a non-uniform circular motion around the center. 
Let $\vec{r}$ be the offset of the sensor from the center of rotation and let
$\vec{\omega}$ be its angular velocity. The linear velocity of the sensor
following the circular trajectory is
\begin{align}
  \vec{v}(t) = \frac{\partial \vec{r}}{\partial t} = \vec{\omega}(t) \!\times\! \vec{r}(t) \, .
\end{align}
Taking the derviative with respect to time results in the linear acceleration:
\begin{align}
  \vec{a} & = \frac{\partial \vec{v}}{\partial t}                                                                                                              \\
          & = \frac{\partial \vec{\omega}}{\partial t} \!\times\! \vec{r} \, + \, \vec{\omega} \!\times\! \frac{\partial \vec{r}}{\partial t}                                                \\
          & = \frac{\partial \vec{\omega}}{\partial t} \!\times\! \vec{r} \, + \, \vec{\omega} \!\times\! \left( \vec{\omega} \!\times\! \vec{r} \right) \label{eq:forces}
\end{align}
An accelerometer measures the gravity in negative direction, e.g., a sensor laying flat on a desk measures the upwards force from the desk pushing against gravity.
Thus, we extend~\eqref{eq:forces} by subtracting gravity as measured by the accelerometer, $\vec{a}_g$.
Rearranging for $\vec{a}_g$ yields
\begin{equation}
  \vec{a}_g = \vec{\omega} \!\times\! \left( \vec{\omega} \!\times\! \vec{r} \right) \, + \, \frac{\partial \vec{\omega}}{\partial t} \!\times\! \vec{r} \, - \, \vec{a} \, ,
  \label{eq:compensation}
\end{equation}
where $\vec{a}$ is the measurement of the accelerometer.

\subsection{Developing the Model for Spherical Systems}
The previous Section~\ref{ssec:derivation} only considers the accelerations due
to rotation of the system. This Section will develop the model including the
linear motion of the platform as well, using the example of a spherical mobile
mapping system with radius $R$ that rolls without slippage on a surface with
normal vector $\vec{n}$. In that case, let $\vec{r}$ be the position of the
sensor inside the rolling ball, composed of $\vec{r}_e$, pointing from the
center of the ball to the sensor, and $\vec{r}_c$, the position of the balls
center:
\begin{align}
  \vec{r} = \vec{r}_e + \vec{r}_c \,.
\end{align}
The same considerations from the previous Section~\ref{ssec:derivation} apply to $\vec{r_e}$, i.e.,
\begin{align}
  \vec{v}_e & = \frac{\partial \vec{r}_e}{\partial t} = \vec{\omega} \!\times\! \vec{r}_e \, , \\
  \vec{a}_e & = \frac{\partial \vec{v}_e}{\partial t} = \frac{\partial \vec{\omega}}{\partial t} \!\times\! \vec{r}_e \, + \, \vec{\omega} \!\times\! \left( \vec{\omega} \!\times\! \vec{r}_e \right) \, .
\end{align}
The linear velocity of the balls center, rolling without slippage, is
\begin{align}
  \vec{v}_c = \frac{\partial \vec{r}_c}{\partial t} = \left( R \cdot \vec{\omega} \right) \!\times\! \vec{n} \,.
\end{align}
Taking the derivative with constant ball radius $R$ yields
\begin{align}
  \vec{a}_c = \frac{\partial \vec{v}_c}{\partial t} = ( R \cdot \frac{\partial \vec{\omega}}{\partial t} ) \!\times\! \vec{n} \, + \, \left( R \cdot \vec{\omega} \right) \!\times\! \frac{\partial \vec{n}}{\partial t} \,.
\end{align}
The combined acceleration of the sensor due to linear and angular motion, excluding gravity, consequently is
\begin{align}
  \vec{a} & = \frac{\partial^2 \vec{r}}{\partial t^2} \nonumber \\
          & = \frac{\partial^2 \vec{r}_e}{\partial t^2} + \frac{\partial^2 \vec{r}_c}{\partial t^2} = \vec{a}_e + \vec{a}_c  \\
          & = \vec{\omega} \!\times\! \left( \vec{\omega} \!\times\! \vec{r}_e \right) +  \frac{\partial \vec{\omega}}{\partial t} \!\times\! \vec{r}_e  +  R \frac{\partial \vec{\omega}}{\partial t} \!\times\! \vec{n} +  R \vec{\omega} \!\times\! \frac{\partial \vec{n}}{\partial t}. \nonumber
\end{align}
Including the measured gravity $\vec{a}_g$ as done in Section~\ref{ssec:derivation} and rearranging yields
\begin{equation}
  \vec{a}_g = \vec{\omega} \!\times\! \left( \vec{\omega} \!\times\! \vec{r}_e \right) +  \frac{\partial \vec{\omega}}{\partial t} \!\times\! \vec{r}_e + R \frac{\partial \vec{\omega}}{\partial t} \!\times\! \vec{n} + R \vec{\omega} \!\times\! \frac{\partial \vec{n}}{\partial t} - \vec{a} .
\end{equation}
However, this model requires a good estimate of the normal vector $\vec{n}$ in the local sensor frame, which is subject to future work.

\subsection{IMU Intrinsic Calibration}
To correct the accelerometer data using the angular motion measured by the gyroscopes, both triads need to provide their readings in a shared coordinate system.
This intrinsic calibration step is indispensable for our proposed extrinsic calibration method, which is why we introduce some common notation here.
For further details regarding derivation and calibration residuals, reffer to the original paper~\cite{imutk2014}. 
The model for intrinsic calibration of the IMU triads, taking into account scale, bias, and misalignments between the triad axes with respect to the accelerometers x-axis, is:
\begin{align}
  \vec{a} &= \mathbf{M}_a \cdot \mathbf{S}_a \cdot \left( \hat{\vec{a}} - \vec{b}_a\right), \\
  \vec{\omega} &= \mathbf{M}_\omega \cdot \mathbf{S}_\omega \cdot \left( \hat{\vec{\omega}} - \vec{b}_\omega\right),
\end{align}
where $\hat{\vec{a}}$ and $\hat{\vec{\omega}}$ are the raw sensor readings of the accelerometer and gyroscope respectively, $\vec{b}_a$ and $\vec{b}_\omega$ represent biases, $\mathbf{S}_a$ and $\mathbf{S}_\omega$ are diagonal scale matrices, and $\mathbf{M}_a$ and $\mathbf{M}_\omega$ are misalignment matrices of the form:
\begin{align}
  \mathbf{M}_a &= \begin{bmatrix} 1 & -\alpha_{yz} & \alpha_{zy} \\ 0 & 1 & -\alpha_{zx} \\ 0 & 0 & 1 \end{bmatrix},\,
  \mathbf{M}_\omega = \begin{bmatrix} 1 & -\gamma_{yz} & \gamma_{zy} \\ \gamma_{xz} & 1 & -\gamma_{zx} \\ -\gamma_{xy} & \gamma_{yx} & 1 \end{bmatrix}. \nonumber
\end{align}
\begin{figure}
  \centering
  \includegraphics[width=.75\linewidth]{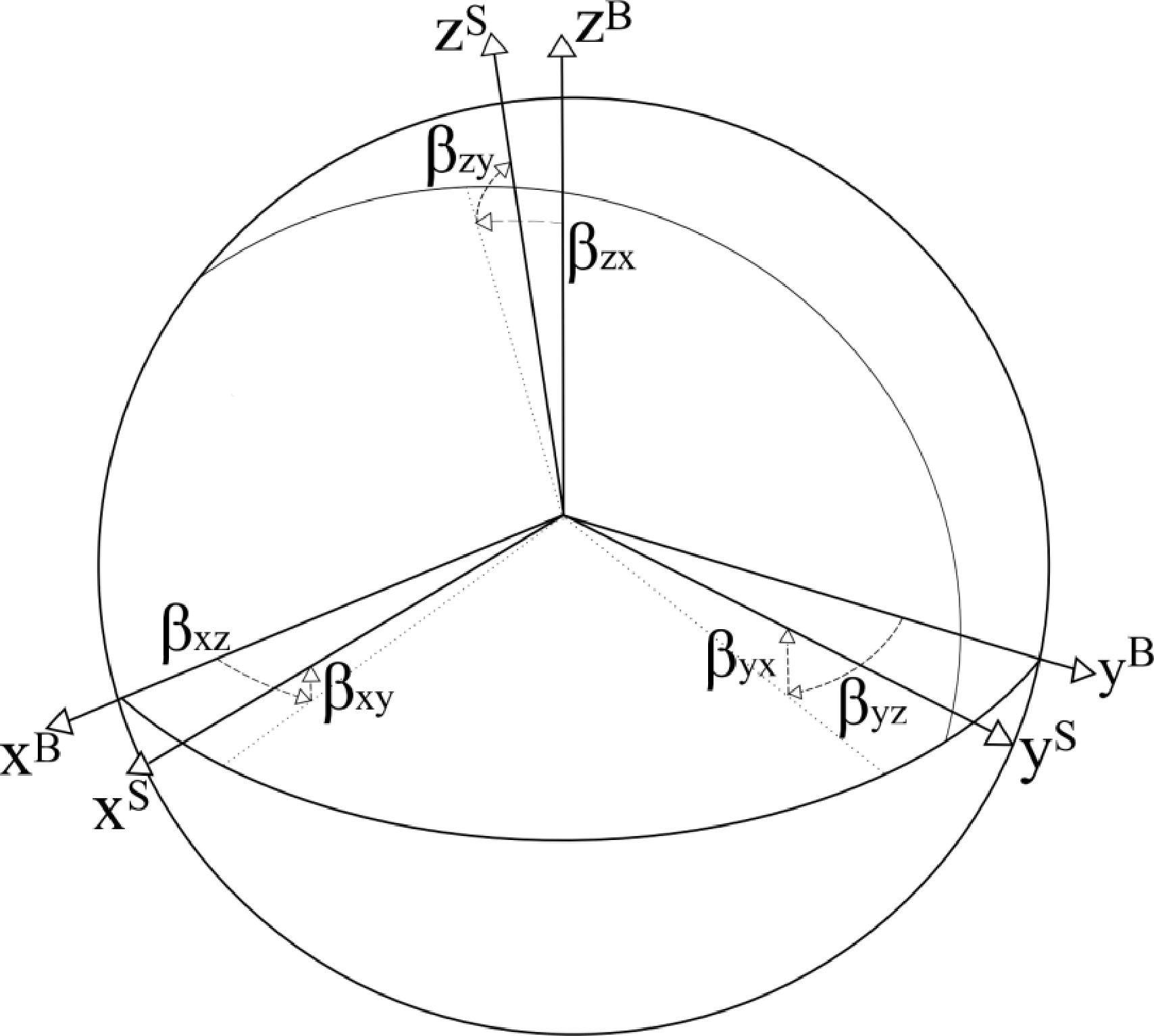}
  \caption{Misalignment angles $\beta$ between an orthogonal body frame $[x^B, y^B, z^B]$ and the non-orthogonal sensor frame $[x^S, y^S, z^S]$. 
  We include the original illustration from~\cite{imutk2014} for completeness.}\label{fig:mis}
\end{figure}
Figure~\ref{fig:mis} illustrates the misalignment angles represented by the matrices. 
We apply the method from~\cite{imutk2014} to calibrate the bias, scale, and misalignment parameters without external equipment, using the multi-orientation approach: We place the IMUs in different orientations for a short duration, then rotate it into another orientation, and repeat.

\subsection{Motion Compensation Calibration}\label{ssec:imucalib}
\begin{figure}
  \centering
  \includegraphics[width=\linewidth]{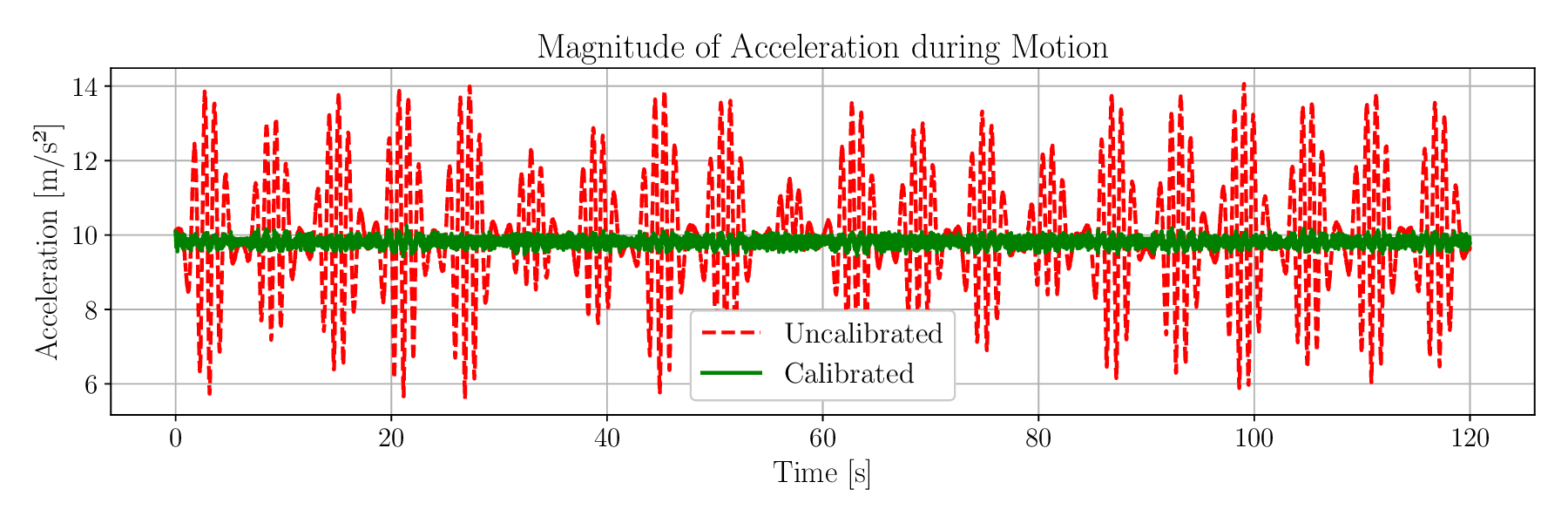}
  \caption{Simulated comparison of total acceleration magnitude before and after the proposed calibration.
  Before calibration (red) motion acceleration effects are present.
  After calibration (green) the measurements are closer to the magnitude of local gravity ($9.81$~m~s$^{-2}$).}\label{fig:magnitude}
\end{figure}
The proposed calibration method is based on the principle that after subtracting the accelerations due to motion from the accelerometer measurements, the result should have the same magnitude as local gravity $\norm*{\vec{g}}^2$, which is illustrated by Figure~\ref{fig:magnitude}.
Thus, to collect an appropriate dataset for calibration, the operator should try to avoid translation of the base when rotating the system, i.e., the center of rotation should stay fixed.
However, the presented method is robust against small deviations from this.
Given $N$ measurements, let $\vec{\omega}_n$ and $\vec{a}_n$ with $n \in \lbrace 1, 2, \dots, N\rbrace$ be corresponding intrinsically calibrated gyroscope and accelerometer measurements.
The optimization problem subject to the error $E \in \mathbb{R}$ is:
\begin{align}
  E &= E(\tilde{\vec{r}}) \nonumber \\
  &= \left(\sum_{n=1}^{N} \abs{ \norm*{\vec{g}}^2 - \norm*{\vec{\omega}_n \!\times\! \left( \vec{\omega}_n \!\times\! \tilde{\vec{r}} \right) + \frac{\partial \vec{\omega}_n}{\partial t} \!\times\! \tilde{\vec{r}} - \vec{a}_n}^2} \right)^2 \nonumber \\
  \vec{r} &= \argmin_{\tilde{\vec{r}}} \left( E(\tilde{\vec{r}}) \right)\,.\label{eq:mindisplacement}
\end{align}
We compute the gradient using automatic differentiation (AD) and minimize Equation~\eqref{eq:mindisplacement} using the well-known Levenberg-Marquardt (LM) optimization algorithm in Ceres.
For AD to work properly, we have to provide a numerically stable implementation of the angular acceleration term $\frac{\partial \vec{\omega}}{\partial t}$.

\subsection{Implementation of Discrete Derivative}
In order to implement the numerical derivative of the gyroscope $\frac{\partial \vec{\omega}}{\partial t}$, which we need during calibration but also during online attitude estimation, we have to consider each axis as a discrete, noisy signal $x[n]$, with $N$ samples and $n \in \lbrace 1,2, \dots, N \rbrace$.
Naivley differentiating numerically using the method of fintie differences leads to amplification of noise, making the result unusable for downstream processing.
Instead, we use a Derivative-of-Gaussian (DoG) kernel. 
A DoG kernel combines smoothing and differentiation in a single convolution.
The Gaussian function is defined as
\begin{align}
G(t) &= \frac{1}{\sigma \sqrt{2\pi}} \exp\!\left(-\frac{t^2}{2\sigma^2}\right),
\end{align}
where $\sigma$ is the standard deviation.
The derivative of the Gaussian is
\begin{align}
\frac{\partial G(t)}{\partial t} &= -\frac{t}{\sigma^2} \, G(t)
= -\frac{t}{\sigma^3 \sqrt{2\pi}} \exp\!\left(-\frac{t^2}{2\sigma^2}\right).
\end{align}
Considering the signal in discrete, variable time steps $\Delta t$, we implement the DoG as a kernel:  
\begin{align}
h_1[k] &= -\frac{t_k}{\sigma^3 \sqrt{2\pi}} 
\exp\!\left(-\frac{t_k^2}{2\sigma^2}\right),
\quad t_k = k \, \Delta t.
\end{align} 
When applying the convolution during calibration, we use the kernel with (odd) size $2K+1$ centered at $k=0$, such that it is symmetric and non-causal:
\begin{align}
\frac{\partial x[n]}{\partial t} &= \sum_{k=-K}^{K} h_1[k] \, x[n-k].\label{eq:kernelnoncausal}
\end{align}
This yields a zero-phase, zero-delay derivative.
For the causal version, which runs online, we apply the same kernel but shifted by $K$ samples, such that it uses only present and past samples:
\begin{align}
  \frac{\partial x[n]}{\partial t} &= \sum_{k=-K}^{K} h_1[k] \, x[n-k-K].
\end{align}
This enforces causality but introduces an inherent delay of $K$ samples. The trade-off is:
Choosing a larger kernel (larger $K$) increases smoothing and delay, while a smaller kernel yields less delay but is noisier.
Thus, the DoG effectivley implements a lowpass filter with a cutoff frequency $f_{\text{cut}}$.
We calculate the standard deviation of the Gaussian $\sigma$ as
\begin{align}
\sigma &= \frac{1}{2\pi f_{\text{cut}}}.
\end{align}
Given the sampling frequency $f_s$ of the sensor, we calculate the kernel window size $W$ (in samples):
\begin{align}
W &= 2K+1 = \left\lceil 6 \,\sigma f_s\right\rceil,
\end{align}
corresponding to approximately $\pm 3\sigma$ of the Gaussian, which represents $99.7\%$ coverage.
\begin{figure}
  \centering
  \includegraphics[width=\linewidth]{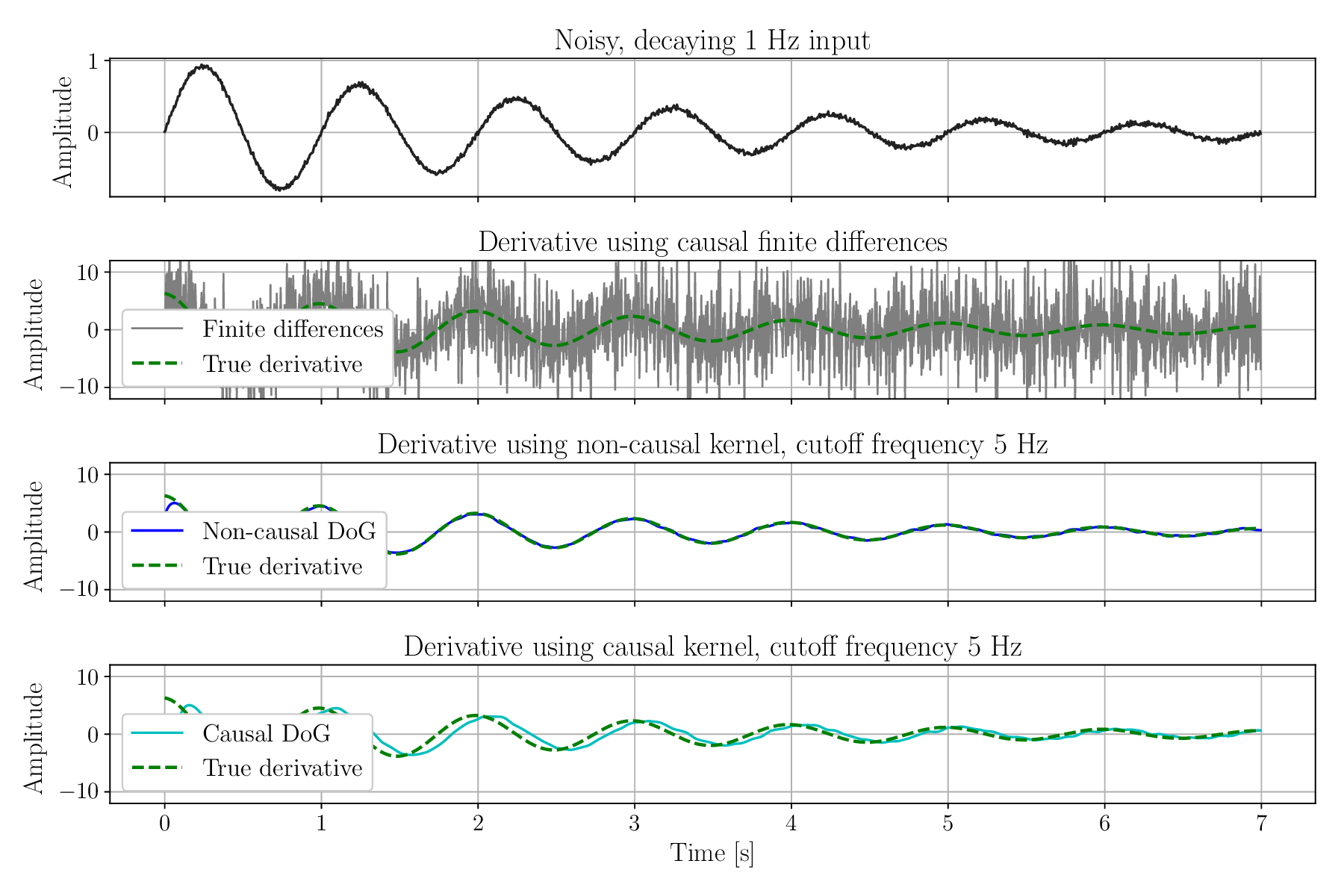}
  \caption{
    Comparison of numerical derivatives on a $1$~Hz sine wave with additional white noise sampled at $200$~Hz.
    In this illustrative example the DoG kernels have a cutoff frequency of $5$~Hz corresponding to a standard deviation of the Gaussian $\sigma \approx 0.032$~s and window size of $N = 39$ samples.
    The method of finite differences amplifies noise, wheras convolution with the DoG kernels produce smooth derivatives.
    In this example, the causal convolution introduces a delay of $0.095$~s of the output, wheras the non-causal convolution does not. 
  }\label{fig:derivatives}
\end{figure}
Figure~\ref{fig:derivatives} illustrates the resulting derivatives of a noisy input signal for the method of finite differences, causal-, and non-causal-DoG kernels.
The non-causal convolution produces a smooth, zero-phase derivative given the selected cutoff frequency of $5$~Hz, which is well suited for the offline calibration procedure.   

To select an appropriate cutoff frequency $f_{\text{cut}}$ for online processing, we consider the phase shift, as well as the signal- and noise-gain, as a function of the cutoff frequency.
\begin{figure}
  \centering
  \includegraphics[width=\linewidth]{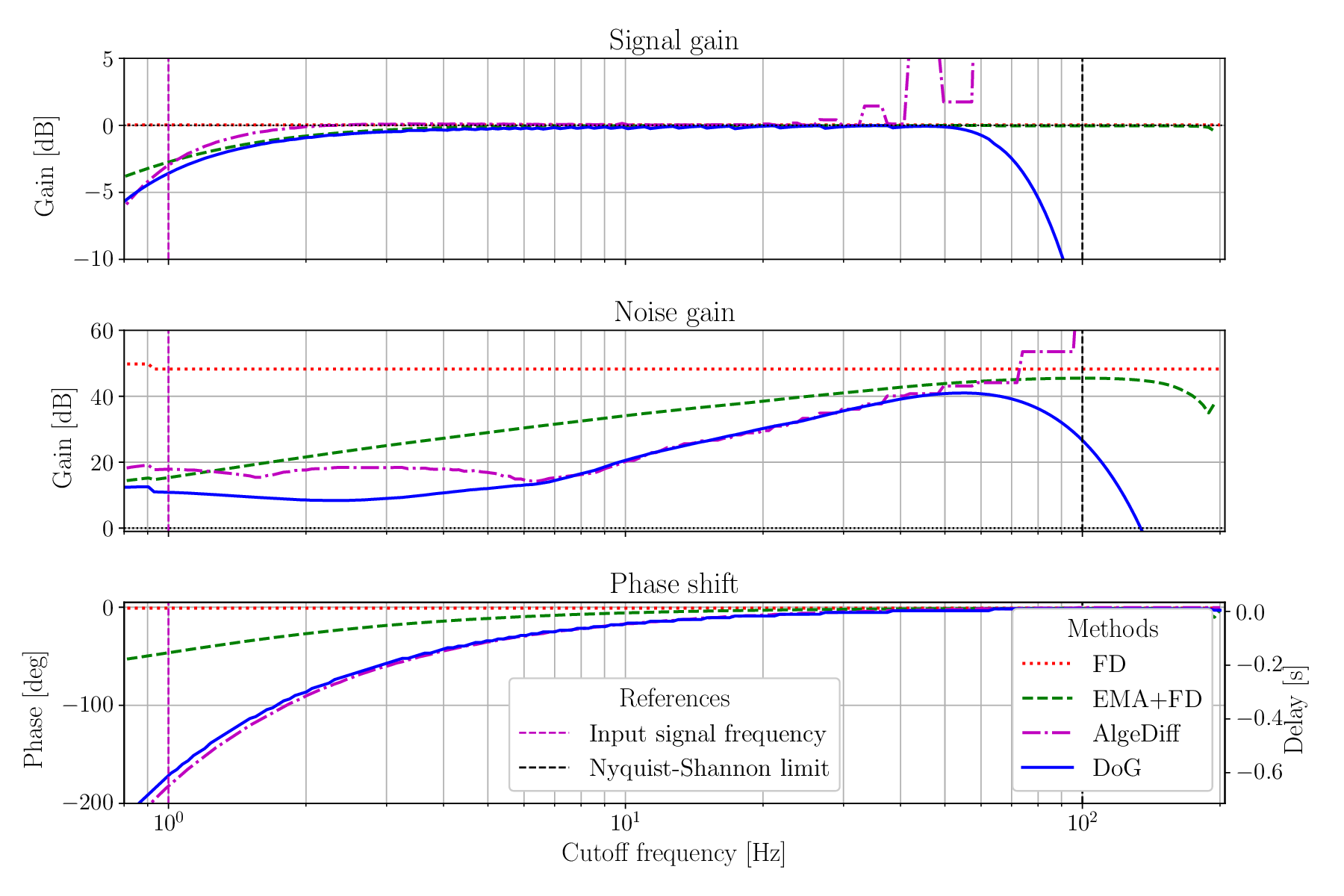}
  \caption{
    Comparison of signal gain, noise gain, and phase shift of causal numerical differentiation methods on a noisy $1$~Hz input signal sampled at $200$~Hz.
    Each method implements smoothing according to a given cutoff frequency to supress the ampliciation of noise.
    The DoG method provides the lowest noise gain across all cutoff frequencies, but also a considerable phase shift.
  }\label{fig:bode}
\end{figure}
Figure~\ref{fig:bode} compares the DoG convolution method with three other alternative methods: 
(1) The method of finite differences (FD), (2) an exponential moving average (EMA) for smoothing, followed by FD, and (3) algebraic differentiators (AlgeDiff) due to~\cite{algediff}.
For a given discrete signal $x$ the EMA produces the smoothed output $y$ by implementing 
\begin{align}
  y[i] = (1 - \alpha) \cdot y[i-1] + \alpha \cdot x[i],
\end{align}
with
\begin{align}
  \alpha &= \sqrt{c^2 - 4c + 3} + c - 1\quad\text{and}\\
  c &= \cos\left(2\pi\cdot\frac{f_{\text{cut}}}{f_{\text{s}}}\right) \nonumber \,,
\end{align}
corresponding to the half-power point where the signal power gain is $10\cdot \log_{10}(0.5) \approx -3$~dB.
We provide a detailed derivation of $\alpha$ in the Appendix~\ref{apx:ema}. 
For the AlgeDiff kernels we use the following parameters: $\alpha = 4.0$, $\beta = 4.0$, $N = 1$, since this results in comparable phase shift and noise gain as the DoG.
However, AlgeDiff produces aliasing effects for high cutoff frequencies, noticable in the signal gain according to Figure~\ref{fig:bode}.    
We select a DoG kernel with $f_{\text{cut}} = 20$~Hz for online processing, since it provides a reasonable tradoff between phase shift and noise gain.

\subsection{Multi-IMU Extrinsic Calibration}\label{ssec:multicalib}

After the intrinsic and extrinsic parameters for each individual IMU have been
found according to Section~\ref{ssec:imucalib}, we now want all IMUs to provide
measurements in the same coordinate system. Thus, in this subsection we will
derive an approach to find for each individual IMU the extrinsic rotation $\mathbf{R}~\in SO(3)$ that transform
their measurements into the coordinate system of the reference IMU.
For the extrinsic calibration we can only use gyroscope data, since the accelerometer merely provides a 2-DoF attitude estimate:
The rotation around the gravity vector can not be recovered.
Thus, we formulate an optimization problem using the intrinsically calibrated gyroscope data.
The resulting rotation then also applies to the accelerometer data since the gyroscope triads are already aligned with the accelerometer triads after intrinsic calibration due to~\cite{imutk2014}.
\newline

\subsubsection{Optimization Problem}

Given $N$ measurements, let $\vec{\omega}_{n}^{i}$ be an intrinsically calibrated measurement of
the $i$-th gyroscope triad, and let $\vec{\omega}_{n}^{\text{ref}}$ be the
corresponding intrinsically calibrated measurement of the reference gyroscope triad, with $n \in \lbrace 1, 2, \dots, N\rbrace$.
We want to find the rotation $\mathbf{R}_{i}^{\text{ref}}~\in SO(3)$ that transforms measurements taken in the $i$-th frame into the reference frame. 
In order to maintain a valid rotation matrix during calibration at all times, we parameterize the optimization residual with the axis-angle $\vec{\theta}~\in~\mathbb{R}^3$, using the capitalized exponential map $\Exp : \mathbb{R}^3 \to SO(3)$ as $\Exp(\vec{\theta}) = \exp(\vec{\theta}^\wedge)$, where $(\cdot)^\wedge$ denotes the skew symmetric operator and $\exp(\cdot)$ is the matrix exponential~\cite{microlie}.
Thus, we construct the optimization problem subject to the error $\vec{e} \in \mathbb{R}^3$ as:
\begin{align}
  \vec{e}\left(\vec{\theta}\right) &= \sum_{n=1}^{N} \norm{\mathbf{R}(\vec{\theta}) \cdot \vec{\omega}_{n}^{i} - \vec{\omega}_{n}^{\text{ref}}}^2 \nonumber \\
  \mathbf{R}_{i}^{\text{ref}} &= \Exp \left({\argmin_{\vec{\theta}} \vec{e}\left(\vec{\theta}\right)}\right)\,,
\end{align} 
which is solved using LM in Ceres.
We then transform the extrinsic vector $\vec{r}_i$ recovered from the motion compensation calibration outlined in Section~\ref{ssec:imucalib} to express it in the reference frame:
\begin{align}
  \vec{r}_i^{\text{ref}} = \mathbf{R}_{i}^{\text{ref}} \cdot \vec{r}_i \,,
\end{align}
which is the vector needed during online motion acceleration compensation.
The full transformation $\mathbf{T} \in SE(3)$ for each IMU describing their offset from the center of rotation and orientation with respect to the reference sensor, is 
\begin{align}
\mathbf{T}_i~=
  \begin{bmatrix}
    \mathbf{R}_{i}^{\text{ref}} & \vec{r}_{i}^{\text{ref}} \\
    \vec{0} & 1
  \end{bmatrix}.
  \label{eq:se3}
\end{align}

\section{Experiments and Evaluation}
We assess the calibration accuracy quantitativley using a Monte-Carlo simulation, which utilizes semi-synthetic IMU data for calibration.
The data is ``semi-synthetic'', because the input angular velocities for the simulation consist of gyroscope data that we record in the real-world.
In the simulation we rigidly mount IMUs in different orientations and positions to a base, which is then rotated according to the input angular velocities.
Thus, we simulate sensor readings for the resulting IMU trajectories.
The implementation of the simulation is outlined in the next Section~\ref{ssec:simulation}.
Additionally, we qualitativley show the impact on the attitude estimation subsystem on real-world data in Section~\ref{ssec:realworld}, using LiDAR data from our spherical mobile mapping system. 

\subsection{Simulation}\label{ssec:simulation}

We implement a simulation that simulates IMU measurements given a 6-DoF trajectory of the sensor.
The trajectory can be fully synthetic, recorded in the real-world, or ``semi-synthetic'', e.g., using real-world angular velocities as input to calculate the sensor trajectory.   
Consider a trajectory as a a set of $N$ discrete poses $\vec{s}[n]$, ordered ascending in $n \in \lbrace 1,\dots,N \rbrace$ according to timestamps $t[n]$, and including the position vector $\vec{p}[n] \in \mathbb{R}^3$ and rotation matrix $\mathbf{R}[n] \in SO(3)$:
\begin{align}
\mathbf{s}[n] &= \lbrace t[n],\; \vec{p}[n],\; \mathbf{R}[n] \rbrace.
\end{align}
If the position and orientation data is not subject to any noise, e.g., if the trajectory is fully synthetic, we use plain central differences to calculate the accelerations due to movement in the world frame $\vec{a}_{m}^{w}$:
\begin{align}
  \Delta t_n &= t[n+1] - t[n-1], \\
  \vec{a}^w_m[n] &= \frac{\partial^2\,\vec{p}[n]}{\partial t^2} \approx \frac{\vec{p}[n+1] - 2 \cdot \vec{p}[n] +\; \vec{p}[n-1]}{\Delta t_{n}^2}.
\end{align}
If the pose data is subject to noise, we instead apply a convolution to each axis with a second order DoG kernel $h_2$:
\begin{align}
t_k[n] &= k \cdot \Delta t_n \\
h_2[k, n] &= \frac{1}{\sigma \sqrt{2\pi}} \left( \frac{t_k^2}{\sigma^4} - \frac{1}{\sigma^2} \right) \exp\!\left( - \frac{t_k^2}{2\sigma^2} \right), \\
\vec{a}^w_{m}[n] &= (h_2 * \vec{p})[n], \label{eq:awm}
\end{align}
where $*$ denotes an axis-wise non-causal convolution, similar to Equation~\eqref{eq:kernelnoncausal}.
In the subsequent evaluation, we select the parameter $\sigma = 15$~ms, corresponding to a width of 3 samples for a $200$~Hz signal.  
The sensor measures an acceleration in its own inertial body frame $\vec{a}_m^b$:
\begin{align}
\vec{a}^{b}_{m}[n] = \vec{R}^{\top}\![n] \cdot \vec{a}^{w}_m[n].\label{eq:abm}
\end{align}
The contribution $\vec{a}^b_g$ due to gravity with magnitude $G$ in the inertial body frame is:
\begin{align}
\vec{g}^{w} &= [0,\;0,\;G]^\top,\\
\vec{a}^{b}_{g}[n] &= \vec{R}^{\top}\![n]\cdot\vec{g}^{w}.\label{eq:abg}
\end{align}
If the positions and orientations of the trajectory contain noise, it propagates through Equation~\eqref{eq:awm},~\eqref{eq:abm}, and~\eqref{eq:abg}.
Thus, the full simulated measurement in the inertial sensor body frame, including signal noise, is
\begin{align}
\mathbf{a}^b[n] &= \mathbf{a}^{b}_{g}[n] + \mathbf{a}^{b}_{m}[n].
\end{align}
If the trajectory is fully synthetic and is not subject to any noise, we simulate accelerometer noise as sampled additive white noise $\vec{\eta}_a$.
Furthermore, we simulate constant sensor bias $\vec{b}_a$, constant scale error $\mathbf{S}_a$ as well as drift $\vec{d}_a$ as Brownian motion: 
\begin{align}
\vec{\eta}_a[n] &\sim \mathcal{N}\!\left(\mathbf{0},\;\frac{\sigma_{a}^2}{\Delta t}\mathbf{I}\right),\nonumber \\
\vec{d}_a[n] &= \sigma_{b_a}\sqrt{\Delta t}\sum_{k=1}^{n}\boldsymbol{\xi}_k,\quad
\boldsymbol{\xi}_k\sim\mathcal{N}(\mathbf{0},\mathbf{I}),\nonumber \\
\vec{a}[n] &= \mathbf{S}_a \cdot \vec{a}^b[n]+\vec{d}_a[n]+\vec{\eta}_a[n]+\vec{b}_a,\label{eq:simacc}
\end{align}
where $\Delta t$ is the mean sampling interval, $\sigma_a$ defines the noise density in units of $\left[\text{m} \cdot \text{s}^{-2} \cdot \text{Hz}^{-0.5}\right]$, and $\sigma_{b_a}$ defines the in-run bias instability in units of $\left[\text{m} \cdot \text{s}^{-3} \cdot \text{Hz}^{-0.5}\right]$.
Analogously, we simulate gyroscope measurements from the orientation sequence.
Given the rotation matrix $\mathbf{R}$ of each pose, we first compute the relative rotation between consecutive samples:
\begin{align}
\Delta \mathbf{R}[n] &= \mathbf{R}^\top[n-1] \cdot \mathbf{R}[n].
\end{align}
Using the capitalized logarithmic map $\Log:SO(3)\rightarrow\mathbb{R}^3$ of the Lie group~\cite{microlie}, the body-frame angular velocity as measured by a gyroscope is
\begin{align}
\boldsymbol{\omega}^b[n] &= \frac{\Log\!\big(\Delta \mathbf{R}[n]\big)}{t[n]-t[n-1]}.\label{eq:liegyro}
\end{align}
Similarly to the accelerometer, we apply an axis-wise convolution with a non-causal Gaussian kernel $h_0$ if the trajectory is not fully synthetic:
\begin{align}
h_0[k, n] &= \frac{1}{\sigma \sqrt{2\pi}} \exp\!\left(-\frac{k^2}{2\sigma^2}\left( t[n]-t[n-1] \right)^2\right),\\
\tilde{\boldsymbol{\omega}}^b[n] &= (h_0 * \boldsymbol{\omega}^b)[n].
\end{align}
If the trajectory is fully synthetic, we do not need smoothing and thus continue with $\tilde{\boldsymbol{\omega}}^b=\boldsymbol{\omega}^b$.
In that case we model constant bias error $\vec{b}_\omega$, constant scale error $\mathbf{S}_\omega$, additive white noise $\vec{\eta}_\omega$, and in-run bias drift $\vec{d}_\omega$:
\begin{align}
\boldsymbol{\eta}_\omega[n] &\sim \mathcal{N}\!\left(\mathbf{0},\;\frac{\sigma_{\omega}^2}{\Delta t}\mathbf{I}\right), \nonumber \\
\mathbf{d}_\omega[n] &= \sigma_{b_\omega}\sqrt{\Delta t}\sum_{k=1}^{n}\boldsymbol{\xi}_k,\quad
\boldsymbol{\xi}_k\sim\mathcal{N}(\mathbf{0},\mathbf{I}),\nonumber \\
\boldsymbol{\omega}[n] &= 
\mathbf{S}_\omega\, \cdot \tilde{\boldsymbol{\omega}}^b[n]+\vec{d}_\omega[n]+\vec{\eta}_\omega[n]+\vec{b}_\omega,\label{eq:simomega}
\end{align}
where $\Delta t$ is the mean sampling interval, $\sigma_{\omega}$ is the gyroscope noise density in units of $\left[\mathrm{rad}\cdot \mathrm{s}^{-1}\cdot \mathrm{Hz}^{-0.5}\right]$,
$\sigma_{b_\omega}$ is the gyroscope in-run bias instability in units of $\left[\mathrm{rad}\cdot \mathrm{s}^{-2}\cdot \mathrm{Hz}^{-0.5}\right]$.
\newline

\subsubsection{Calibration}
To quantify the accuracy of the proposed calibration method we use a Monte-Carlo simulation that calibrates $2000$ IMUs using semi-synthetic data.
We generate the input trajectories for the semi-synthetic datasets by first recording real-world gyroscope data
\begin{align}
  \boldsymbol{\omega}_{\mathrm{in}}[n] &= \lbrace t[n],\;\hat{\vec{\omega}}[n]\rbrace,
\end{align}
with timestamps $t[n]$ and measured angular velocities $\hat{\vec{\omega}}[n]$.
In the simulation, we rigidly mount multiple sensors to a rotating base. 
Let the sensor rigid transformations $\mathbf{T}_i \in SE(3)$ according to Equation~\eqref{eq:se3} be known:
\begin{align}
  \mathbf{T}_i = 
  \begin{bmatrix}
    \mathbf{R}_i & \vec{t}_i \\
    \vec{0} & 1
  \end{bmatrix}.
\end{align}
In our implementation, the reference orientation in the inertial frame is always the identity $\mathbf{R}_{\text{ref}} = \mathbf{I}$, such that its initial orientation coincides with the world frame, and $\mathbf{R}^{\text{ref}}_i = \mathbf{R}_i$.
We use discrete integration of the input angular velocity to obtain the orientation $\mathbf{R}_b$ of the base
\begin{align}
  \mathbf{R}_b[n] &= \mathbf{R}_b[n-1] \cdot \Exp\!\big(\boldsymbol{\hat{\omega}}[n] \cdot \left(t[n]-t[n-1]\right) \big) .
\end{align}
Thus, the full semi-synthetic sensor trajectories $\vec{s}_i$ in world coordinates, serving as the input for the simulation, are
\begin{align}
  \bar{\vec{p}}_i[n] &= \mathbf{R}_b[n] \cdot \mathbf{t}_{i} \;,\\
  \bar{\mathbf{R}}_i[n] &= \mathbf{R}_b[n] \cdot \mathbf{R}_{i} \;,\\
  \mathbf{s}_{i}[n] &= \lbrace t[n],\;\bar{\vec{p}}_i[n],\;\bar{\mathbf{R}}_i[n] \rbrace.
\end{align}
\begin{figure*}
  \centering
  \includegraphics[width=\linewidth]{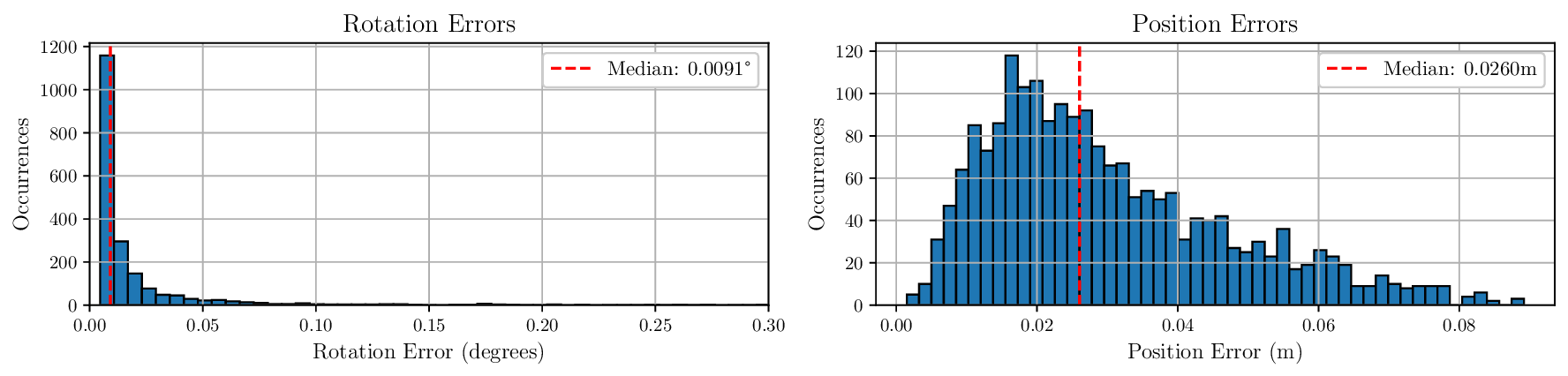}
  \caption{Histograms that show the distribution and median of calibration RMSE in the Monte-Carlo simulation that simulates $2000$ sensor calibrations with semi-synthetic data. 
  (Left:) Rotation error. The ground truth rotations have been sampled uniformly across Euler angles. 
  (Right:) Position error. The ground truth position vectors $\vec{p} = [p_x, p_y, p_z]$ have been sampled uniformly across the interval $ p_x, p_y, p_z \in [-0.5, 0.5]$~m.}\label{fig:histograms}
\end{figure*}
The simulation samples the ground truth offset vectors $\vec{t}_i$ uniformly from a cube with $1$~m side length, and the ground truth rotations are uniformly sampled in Euler coordinates. 
Figure~\ref{fig:histograms} shows the resulting distribution of rotation and position errors.
The distributions show a median position error of $2.6$~cm and rotation error of $0.0091^{\circ}$, indicating the accuracy to which the method recovers the known rigid body transformation $\mathbf{T}_i$ of each sensor. 
However, these estimates are likely influenced by the input angular velocities used to create the semi-synthetic data and thus, might not reflect the true accuracy of the proposed method.
We want to explore other possible dependencies and test the method with different hardware in future work.
\newline

\subsubsection{Attitude Estimation}
\begin{figure*}
  \centering
  \includegraphics[width=.29\linewidth]{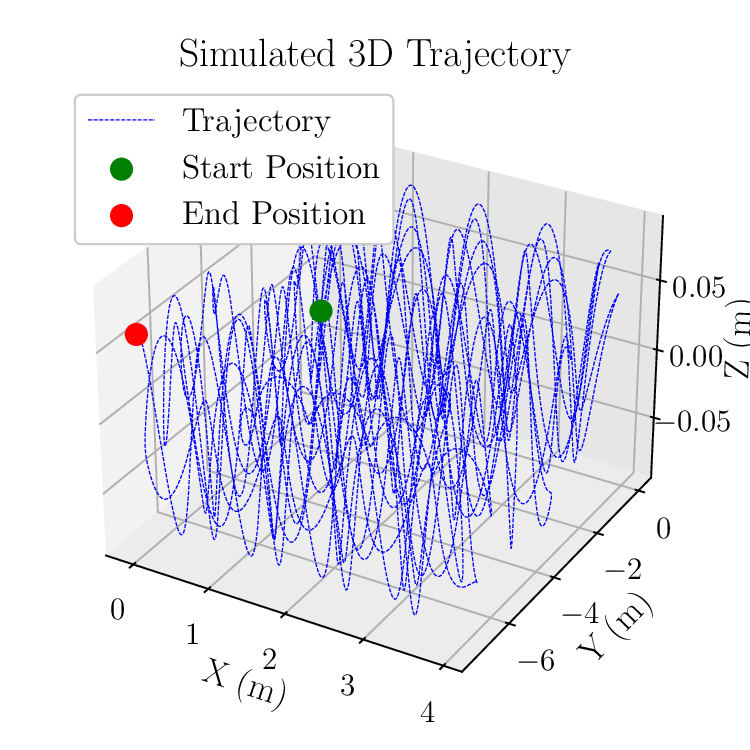}
  \includegraphics[width=.69\linewidth]{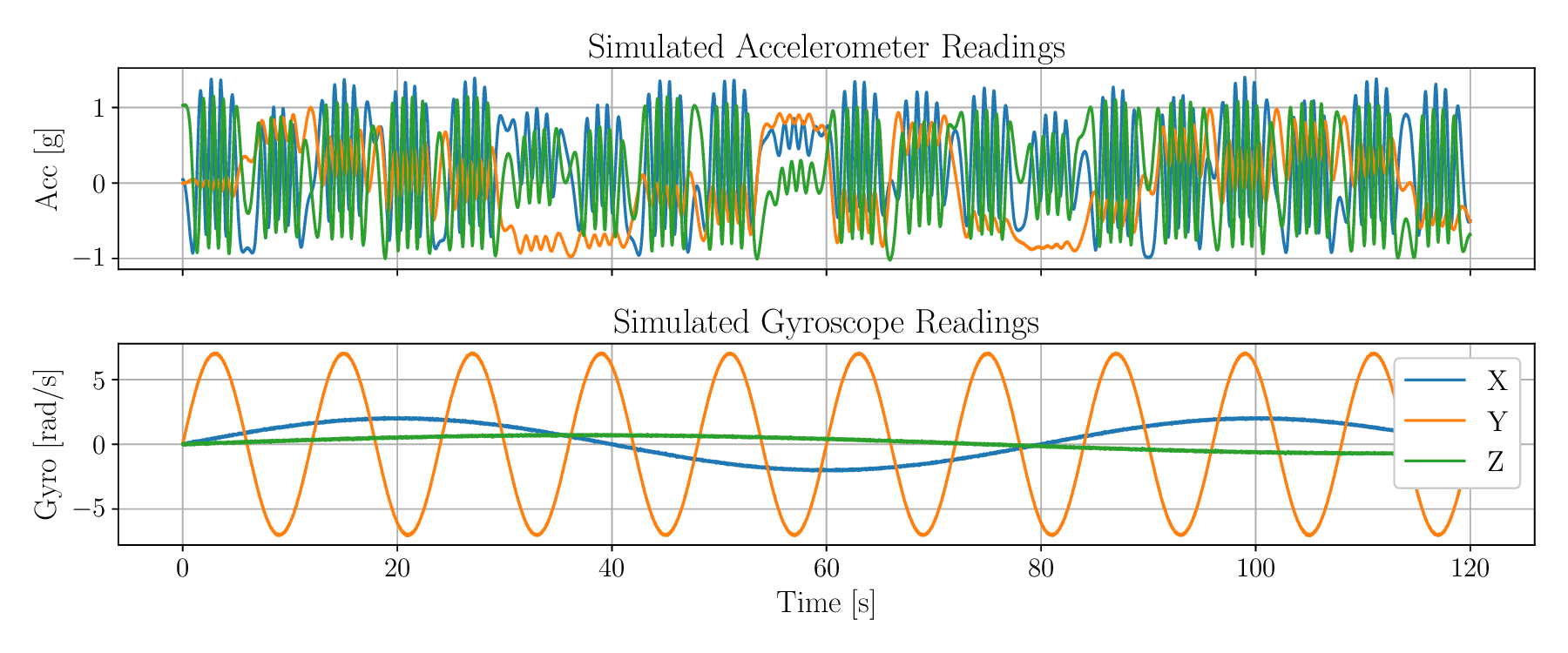}
  \caption{Example of a simulated fully-synthetic dataset used for comparison of attitude estimation algorithms.
  (Left:) Trochoidal trajectory generated from true angular velocities via the trochoidal motion model due to~\cite{trochoidal}.
  (Right:) Corresponding simulated accelerometer and gyroscope measurements, subject to bias, noise, drift, and scale errors.
  Simulated parameters: Accelerometer noise density $\sigma_a = 0.002~\left[\text{m} \cdot \text{s}^{-2} \cdot \text{Hz}^{-0.5}\right]$, 
  accelerometer in-run bias instability $\sigma_{b_a} = 0.00002~\left[\text{m} \cdot \text{s}^{-3} \cdot \text{Hz}^{-0.5}\right]$,
  gyroscope noise density $\sigma_{\omega} = 0.002~\left[\mathrm{rad}\cdot \mathrm{s}^{-1}\cdot \mathrm{Hz}^{-0.5}\right]$,
  and gyroscope in-run bias instability $\sigma_{b_\omega} = 0.0002~\left[\mathrm{rad}\cdot \mathrm{s}^{-2}\cdot \mathrm{Hz}^{-0.5}\right]$.
  In this example, we do not simulate scale and bias errors, since attitude estimation happens after applying calibration.}\label{fig:simtraj}
\end{figure*}

\begin{figure*}
  \centering
  \includegraphics[height=0.275\textheight]{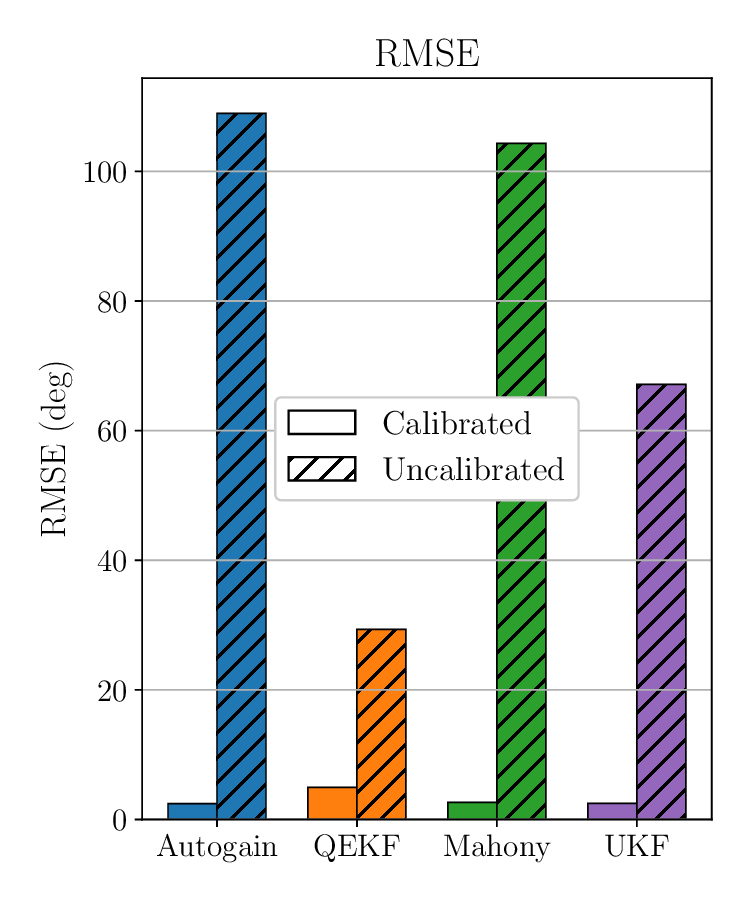}\hfill
  \includegraphics[height=0.275\textheight]{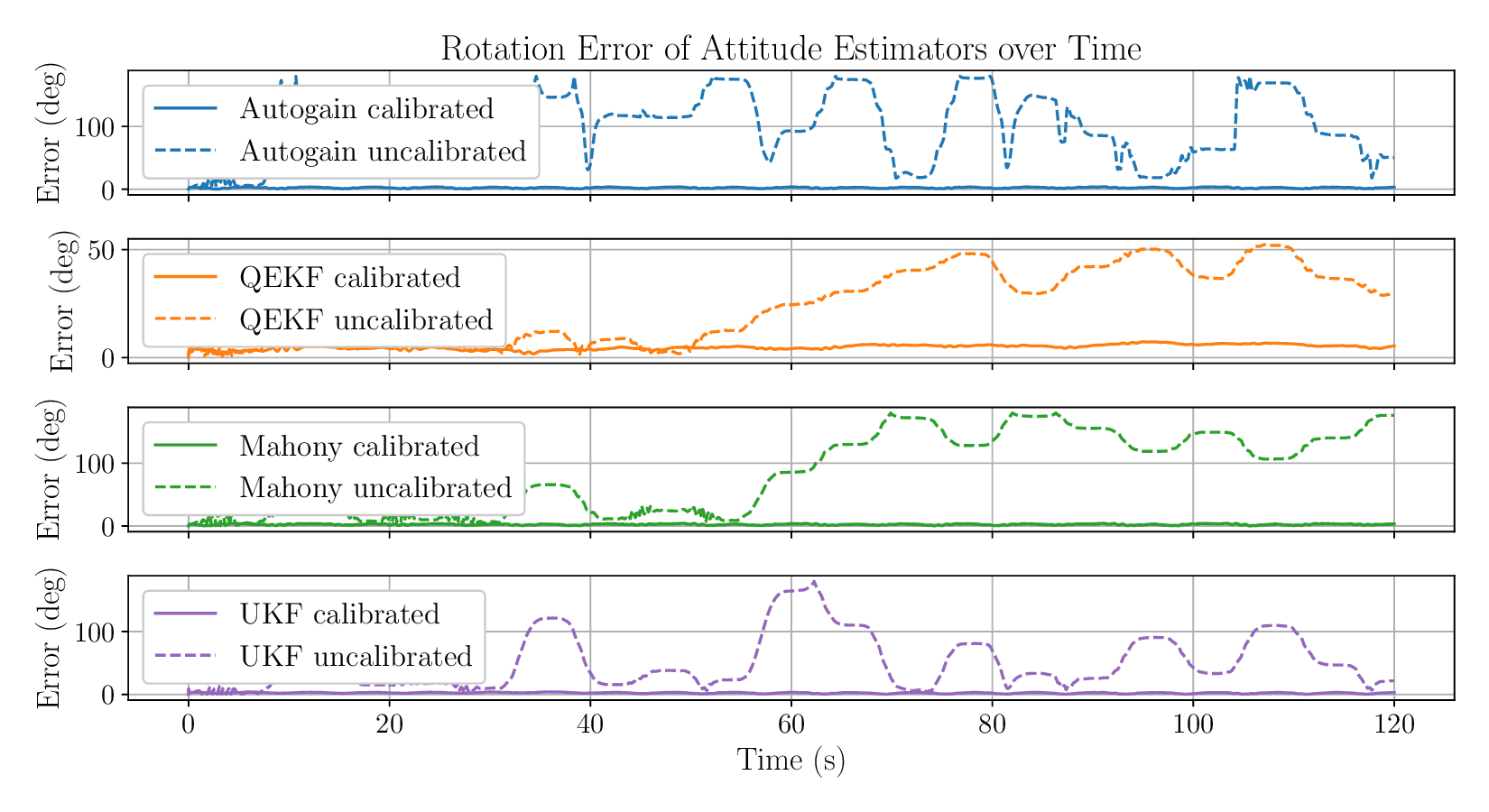}\hfill
  \caption{Comparison of rotation errors on attitude estimators with and without motion-acceleration compensation.
  Input to the estimators in the illustrated example are the fully-synthetic simulated accelerometer and gyroscope readings from Figure~\ref{fig:simtraj}.
  We use the following parameters for each estimator. 
  Autogain~\cite{imujasper}: $\Theta = 0.1$, $\theta = 0.1$.
  QEKF~\cite{qekf}: $P_{\text{init}} = 1$, $\sigma_a = 0.005$, $\sigma_g = 0.005$.
  Mahony~\cite{mahony}: $K_p = 50$, $K_i = 0.01$.
  (Left:) Root mean squared of rotation errors. 
  (Right:) Rotation errors as timeseries.}\label{fig:simerror}
\end{figure*}

In this section we consider the influence of the proposed calibration method on popular downstream attitude estimation algorithms, using fully synthetic data.    
The simulation first creates true angular velocities which are input to a trochoidal motion model~\cite{trochoidal}, resembling the trajectory of a point rigidly mounted inside a rolling ball.
We use the resulting trochoidal sensor trajectories to simulate measurements according to Equations~\eqref{eq:simacc} and~\eqref{eq:simomega}.
Figure~\ref{fig:simtraj} shows an example of the simulated trochoidal trajectory, as well as the corresponding sensor measurements.
These simulated measurements are input to three state-of-the-art attitude estimation algorithms: 
(1) The ``Autogain'' filter which is specialized for spherical systems~\cite{imujasper}, combining the well-known Madgwick- with the Complementary-filter~\cite{madgwick,complementary}. 
(2) A quaternion-based Extended Kalman Filter (QEKF)~\cite{qekf}.
We use the magnetometer-free reference implementation from the AHRS toolbox\footnote{\url{https://ahrs.readthedocs.io/en/latest/filters/ekf.html}}.
Finally, (3) the Mahony filter~\cite{mahony}.
Instead of the raw accelerometer data, each estimator processes the compensated data due to Equation~\eqref{eq:compensation}, using the calibrated offset vector.
Figure~\ref{fig:simerror} shows the resulting attitude estimtation errors for each algorithm with and without applying the proposed method.
The errors of each estimator significantly reduce after applying calibration and motion compensation to the accelerometer data.
Without compensation, the attitude estimators accumulate error quickly due to the high angular velocity, such that both noisy gyroscope integration, as well as gravity measurements from the accelerometer, are unreliable. 

\subsection{Real-World}\label{ssec:realworld}

We test the proposed calibration and compensation approach on our spherical mobile mapping system.
\begin{figure}
  \centering
  \includegraphics[height=0.17\textheight]{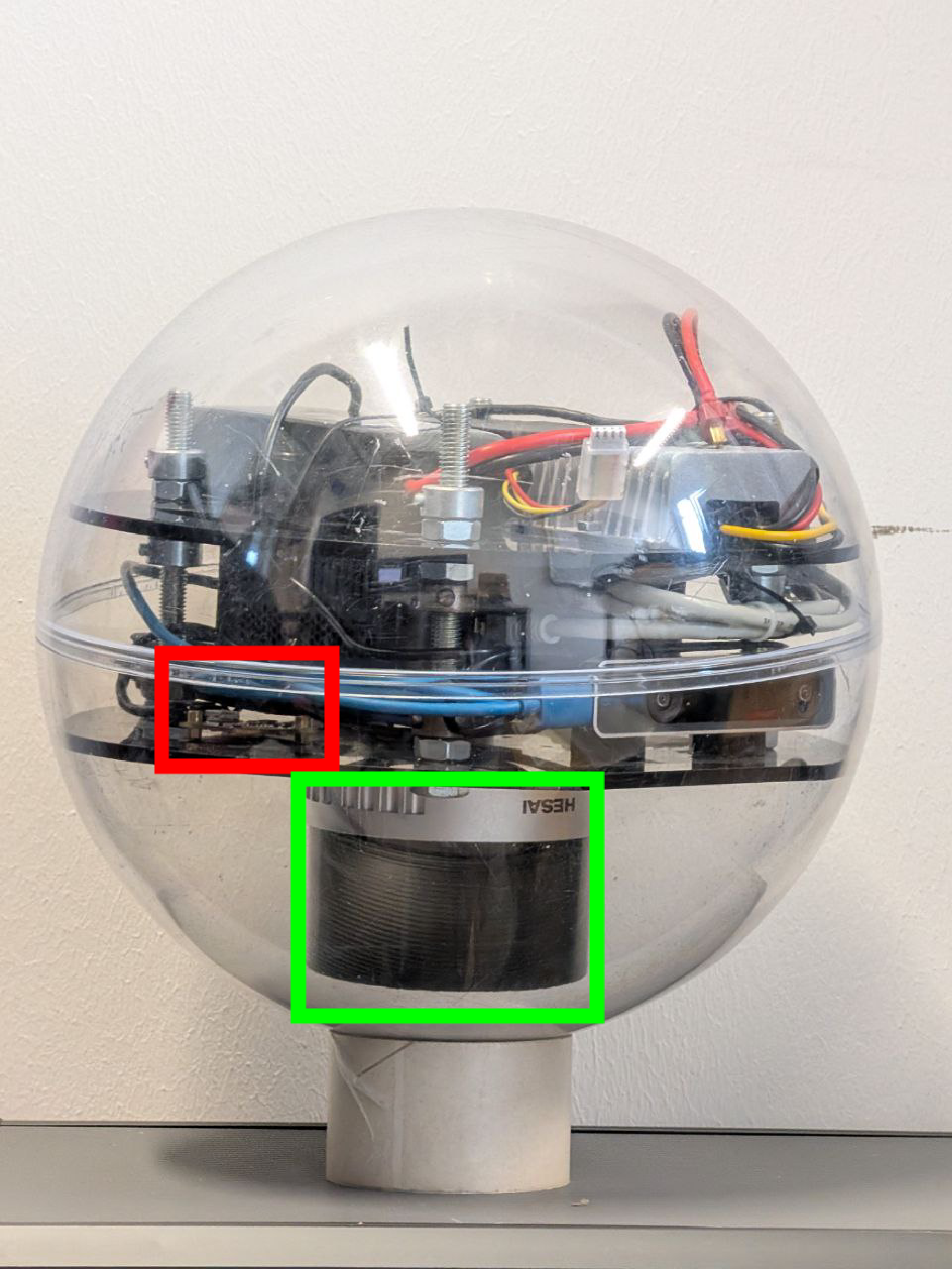}
  \includegraphics[height=0.17\textheight]{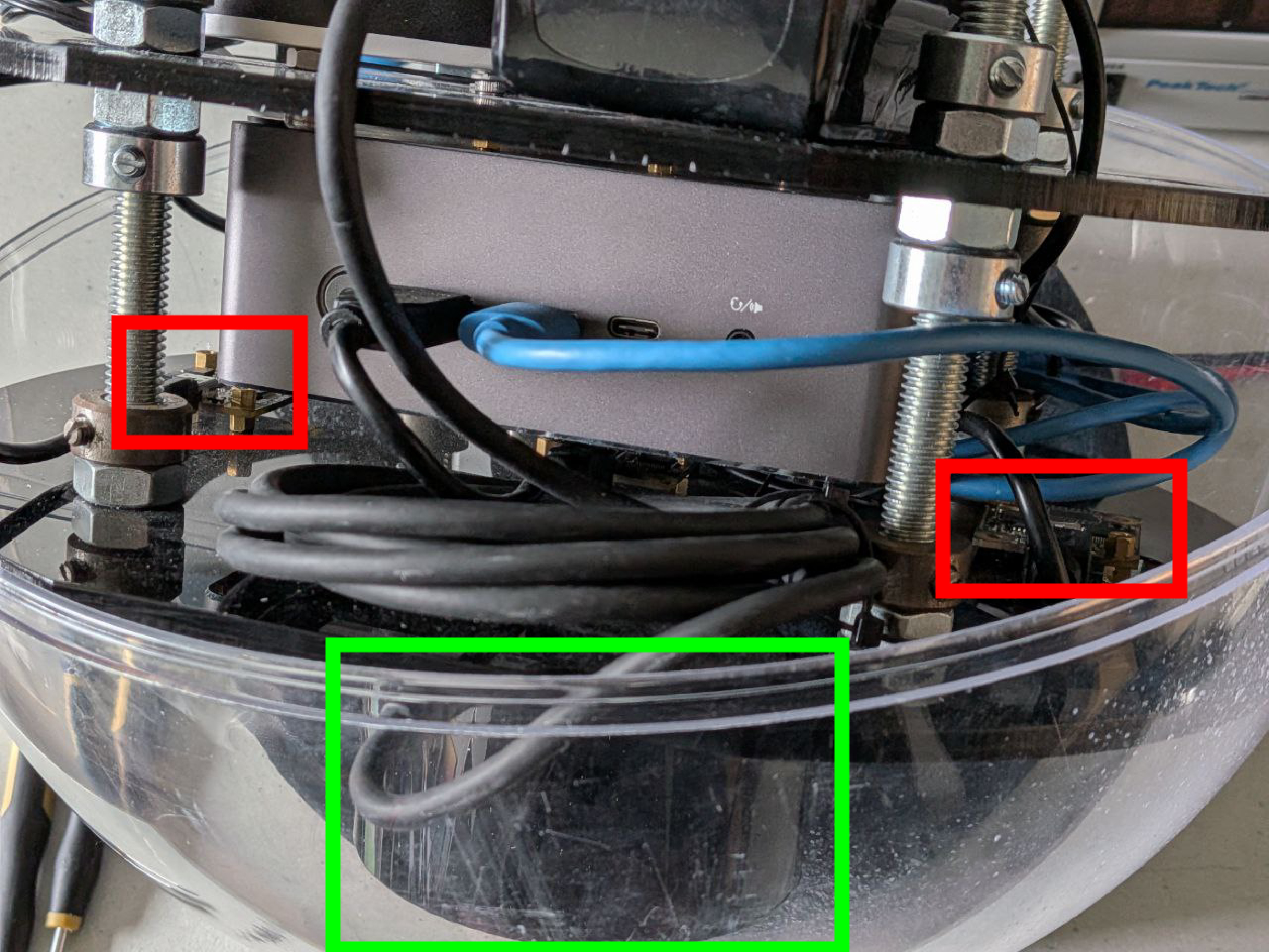}
  \caption{The spherical mobile mapping system from our lab. 
  It is equipped with symmetrically placed Phidgets Spatial 3/3/3 IMUs (marked red), as well as a Hesai Pandar-XT32 laser scanner (marked green).}\label{fig:sphere}
\end{figure}
\begin{figure}
  \centering
  \includegraphics[width=0.32\linewidth]{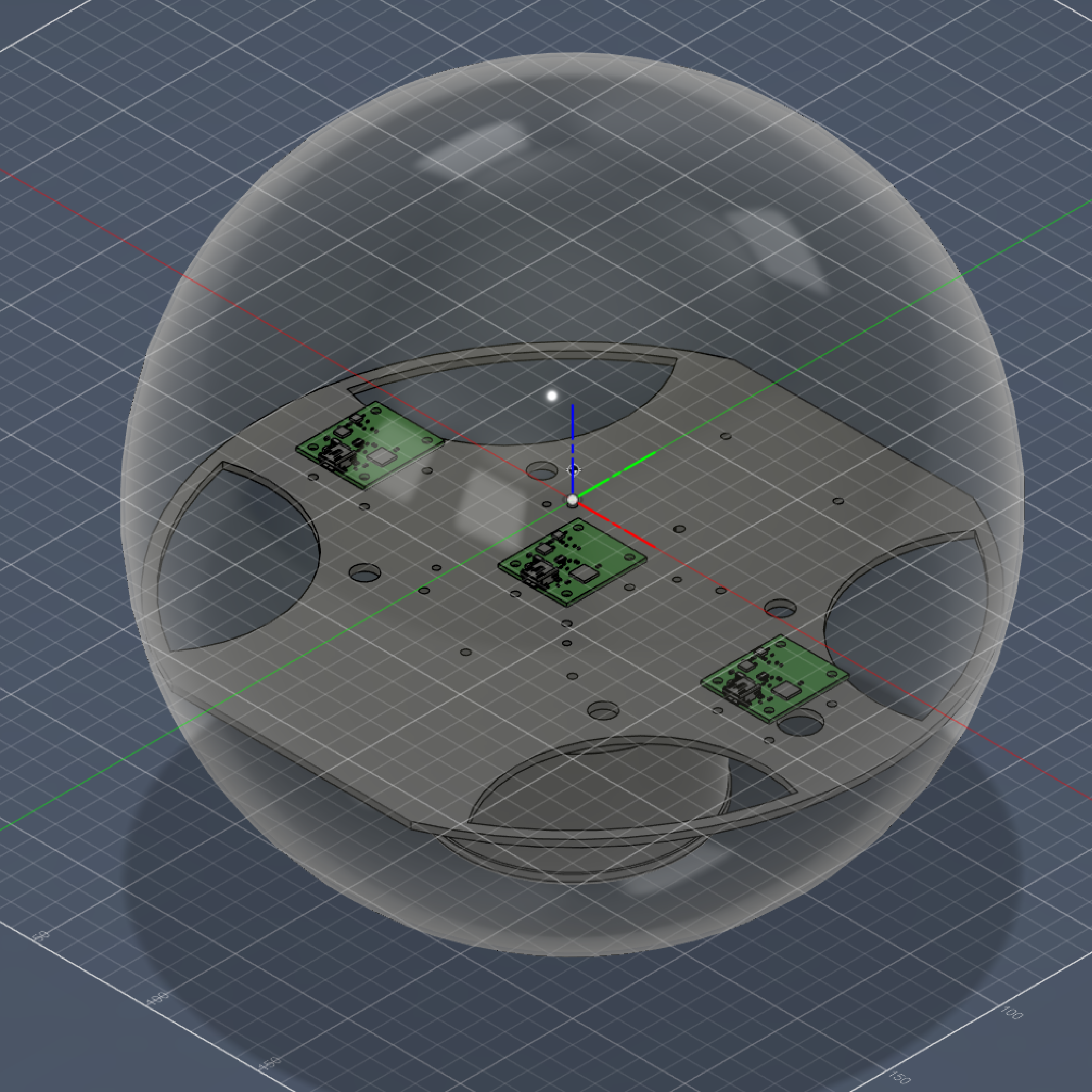}
  \includegraphics[width=0.32\linewidth]{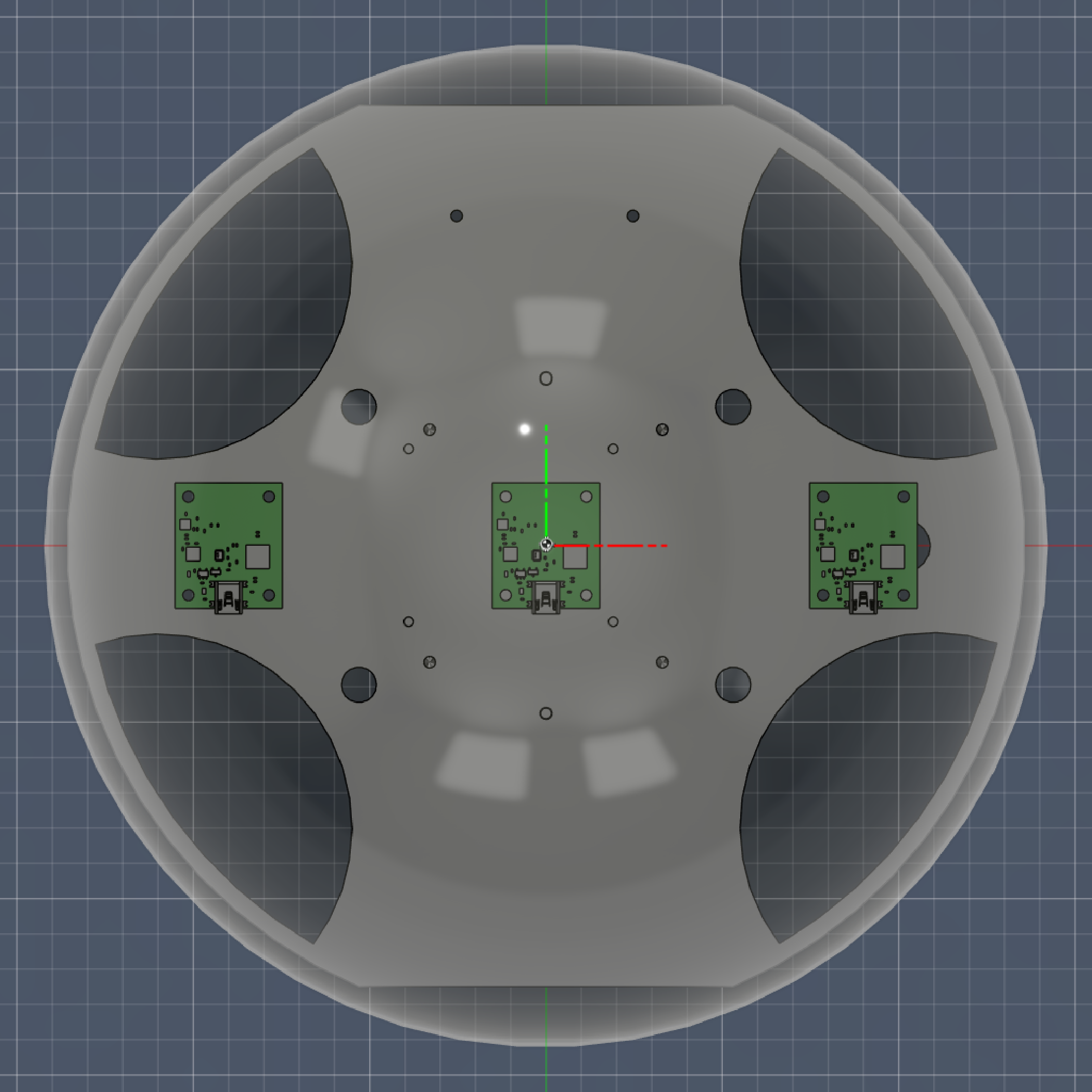}
  \includegraphics[width=0.32\linewidth]{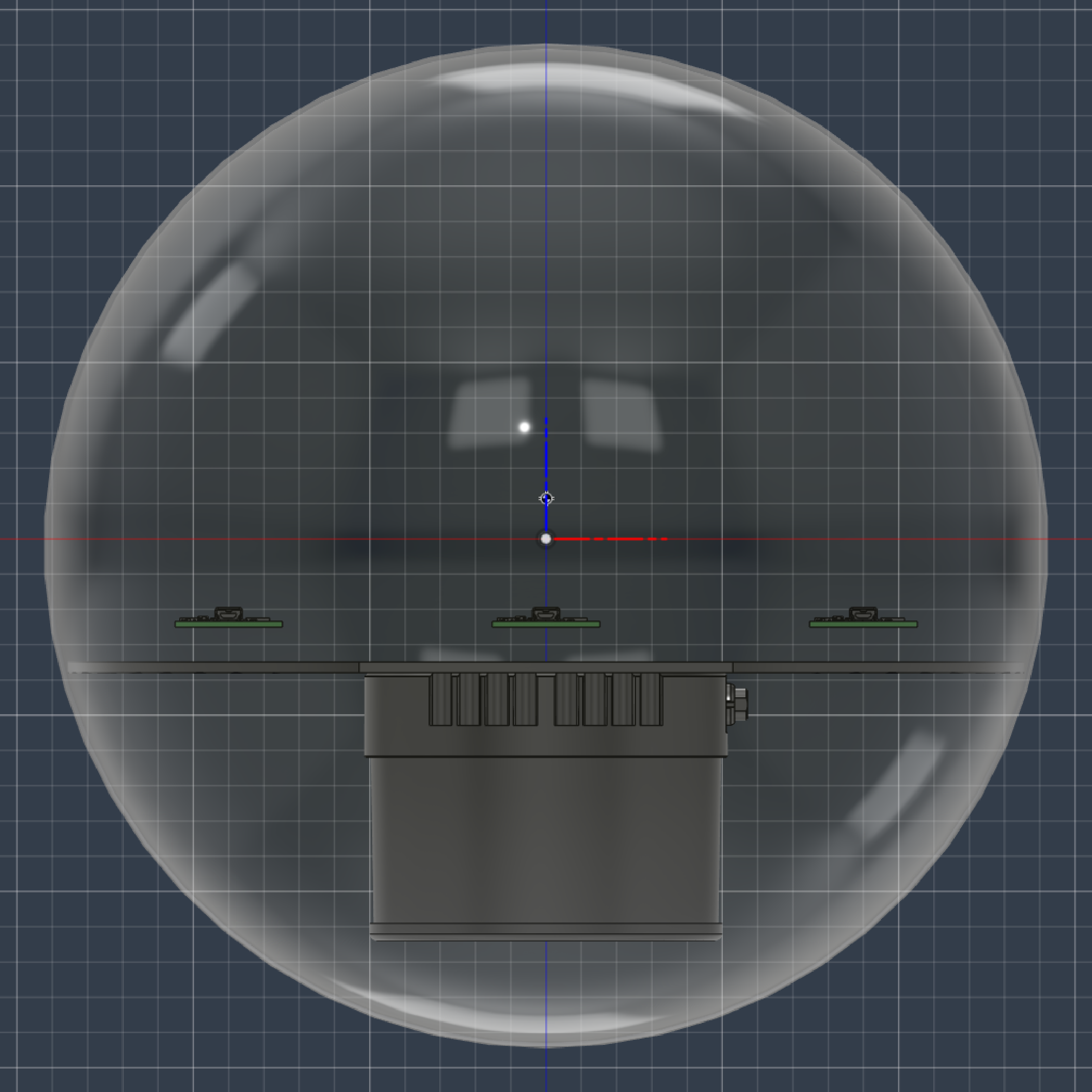}
  \caption{Partial CAD model of the spherical mobile mapping system, highlighting the placement of the IMUs.
  Other parts are invisible except for the shell, laser scanner, and IMUs.
  (Left:) Isometric view.
  (Center:) Orthographic view from above.
  (Right:) Orthographic view from the front.}\label{fig:spherecad}
\end{figure}
Figure~\ref{fig:sphere}~and~\ref{fig:spherecad} shows the prototype from our lab.
On-board sensors includes symmetrically placed Phidget Spatial 3/3/3 1044b IMUs and a Hesai Pandar-XT32 laserscanner.
During the data collection phase for the calibration we place the spherical system on a roll of tape to ensure stability in the static periods.
Since a roll of tape is a common item, we do not consider it as additional equipment.     
The on-board software running on the prototype, including the implementation of motion acceleration compensation, is available at our labs GitHub.\footnote{https://github.com/JMUWRobotics/sphere-mobile-mapping}
\begin{table}[]
  \caption{Comparison of IMU positions obtained from the CAD model and the proposed calibration.}\label{tab:imus}
  \centering
  \resizebox{\linewidth}{!}{%
    \begin{tabular}{rccc|ccc|ccc}
    \multicolumn{1}{l}{}                      & \multicolumn{3}{c|}{IMU0}                                    & \multicolumn{3}{c|}{IMU1}                                    & \multicolumn{3}{c}{IMU2}                                                            \\ \cline{2-10} 
    \multicolumn{1}{l|}{}                     & \multicolumn{1}{c|}{x}    & \multicolumn{1}{c|}{y}    & z    & \multicolumn{1}{c|}{x}    & \multicolumn{1}{c|}{y}    & z    & \multicolumn{1}{c|}{x}     & \multicolumn{1}{c|}{y}     & \multicolumn{1}{c|}{z}    \\ \hline
    \multicolumn{1}{r|}{CAD {[}cm{]}}         & \multicolumn{1}{c|}{9.00} & \multicolumn{1}{c|}{0.00} & 2.35 & \multicolumn{1}{c|}{0.00} & \multicolumn{1}{c|}{0.00} & 2.35 & \multicolumn{1}{c|}{-9.00} & \multicolumn{1}{c|}{0.00}  & \multicolumn{1}{c|}{2.35} \\ \hline
    \multicolumn{1}{r|}{Calibration {[}cm{]}} & \multicolumn{1}{c|}{7.68} & \multicolumn{1}{c|}{0.19} & 2.95 & \multicolumn{1}{c|}{0.33} & \multicolumn{1}{c|}{0.04} & 2.88 & \multicolumn{1}{c|}{-7.21} & \multicolumn{1}{c|}{-0.38} & \multicolumn{1}{c|}{3.22} \\ \hline
    \multicolumn{1}{r|}{Difference {[}cm{]}}  & \multicolumn{3}{c|}{1.46}                                    & \multicolumn{3}{c|}{0.63}                                    & \multicolumn{3}{c}{2.03}                                                           
    \end{tabular}
  }
\end{table}
Table~\ref{tab:imus} compares the resulting extrinsic vector from the center of rotation to the sensor for each on-board IMU.
Additionally, we consider four different cases to evaluate the effectivenes of the proposed calibration and compensation software qualitativley:
(a) Using a single non-centered IMU, performing only intrinsic calibration due to~\cite{imutk2014} and no motion compensation, 
(b) a single non-centered IMU, performing the proposed calibration and motion compensation, 
(c) a single centered IMU, performing only intrinsic calibraiton without motion compensation,
(d) a single centered IMU, performing performing the proposed calibration and motion compensation,
(e) multiple symetrically placed non-centered IMUs, performing only intrinsic calibration without motion compensation, and
(f) multiple symetrically placed non-centered IMUs, performing the proposed calibration and motion compensation.
The placement of the IMU in (c) and (d) is only approximate centered, but closer as in (a) and (b), c.f.,~Figure~\ref{fig:spherecad}.
In cases (e) and (f) we average the measurements of multiple IMUs.
In the latter case the averaging happens after applying the intrinsic calibration and compensating the motion accelerations.
Since the sensors are placed symetrically around the center of rotation, the motion accelerations theoretically cancel when averaging.
\begin{figure*}
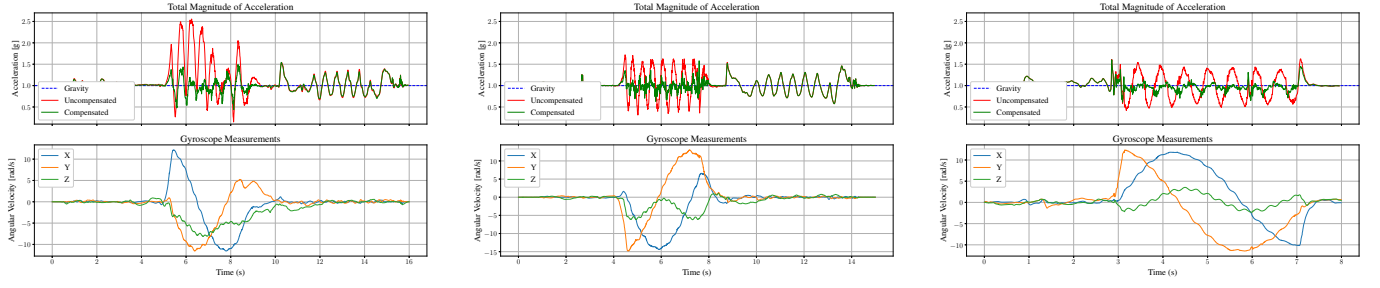

  \centering
  \resizebox{.32\textwidth}{!}{\input{img/real_compensation_single.pgf}}\hfill
  \resizebox{.32\textwidth}{!}{\input{img/real_compensation_single_center.pgf}}\hfill
  \resizebox{.32\textwidth}{!}{\input{img/real_compensation.pgf}}
  \caption{Comparison of compensated and uncompensated real-world IMU data from our spherical mobile mapping system.
  (Left:) Single off-centered IMU.
  (Center:) Single centered IMU.
  (Right:) Average of multiple, symetrically placed IMUs.
  }\label{fig:realdata}
\end{figure*}
Figure~\ref{fig:realdata} shows three example datasets.
The left column corresponds to case (a) and (b) and plots data from a single off-centered IMU.
The column in the center corresponds to case (c) and (d) and plots data from a single centered IMU.
The right column corresponds to case (e) and (f) and plots averaged data from multiple symmetric IMUs.
The above row shows the magnitude of measured acceleration where red corresponds to no compensation and green corresponds to active compensation.
The compensated accelerations are closer to the magnitude of local gravity for both datasets.
This is an expected result, similar to Figure~\ref{fig:magnitude} which shows simulated data corresponding to the trajectory in Figure~\ref{fig:simtraj}.
Although the magnitude of angular velocity in the real-world dataset is similar to the simulated one, the compensated accelerometer measurements in the real-world deviate more from the magnitude of local gravity than in the simulation.  
We are uncertain why this happens but contribute it to non-linearities in the accelerometer sensor which we did not model in the simulation nor in the calibration.
If that is true, it implies that the direction of the gravity vector is actually correct, it is just scaled improperly.
Another way to evaluate the effectivenes of the proposed method qualitativley is to use the compensated and uncompensated accelerometer measurements in an attitude filter and apply the resulting estimations to point data available from the on-board LiDAR.
The filter we use is~\cite{imujasper} since it also estimates the position of the spherical system when rolling on a flat surface without slippage. 
\begin{figure*}
  \centering
  \begin{minipage}[t]{0.485\textwidth}
    \centering
    \subfloat[]{%
      \begin{minipage}[b]{0.485\linewidth}
        \centering
        \includegraphics[width=\linewidth]{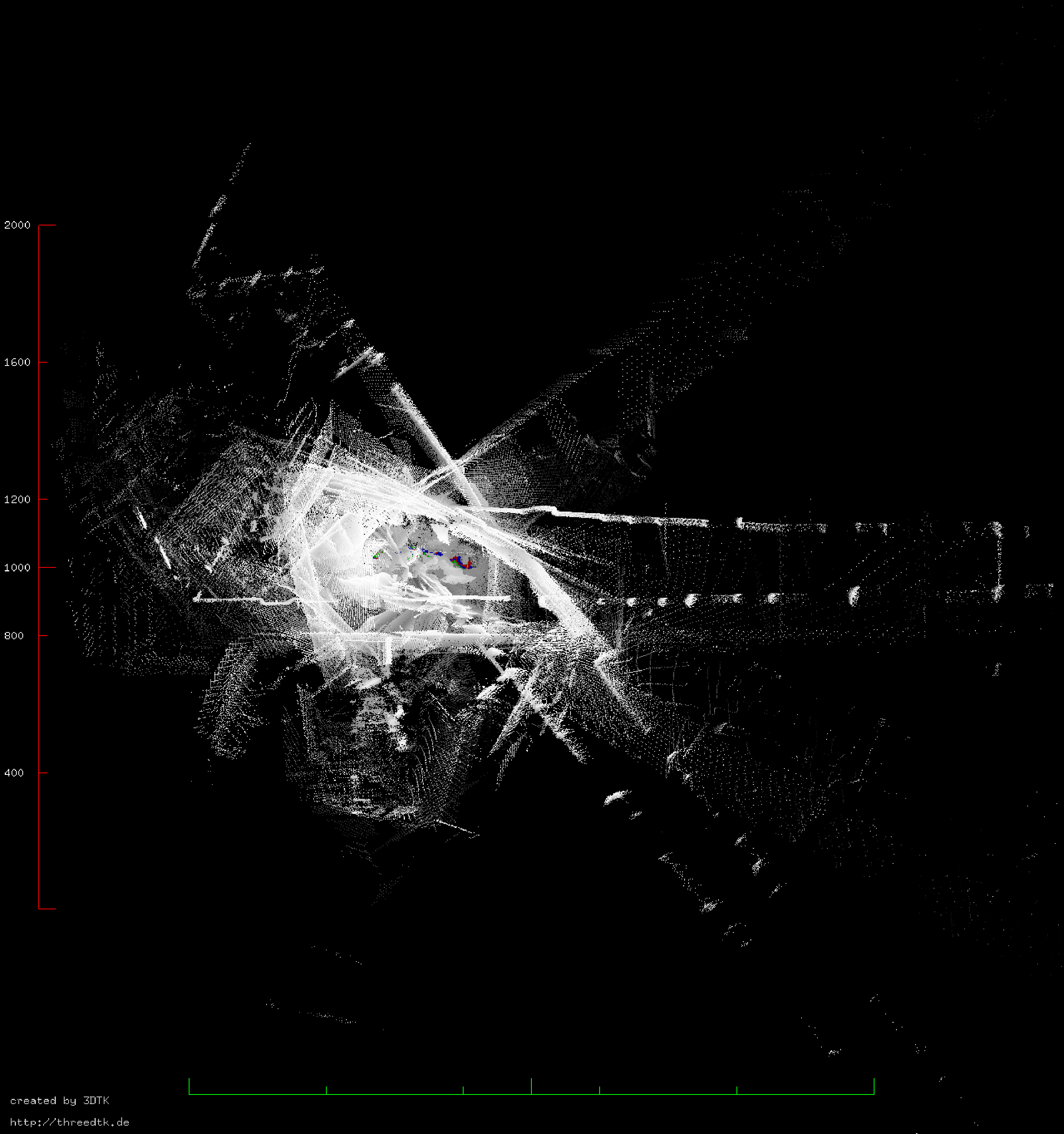}\vspace{1pt}
        \includegraphics[width=\linewidth]{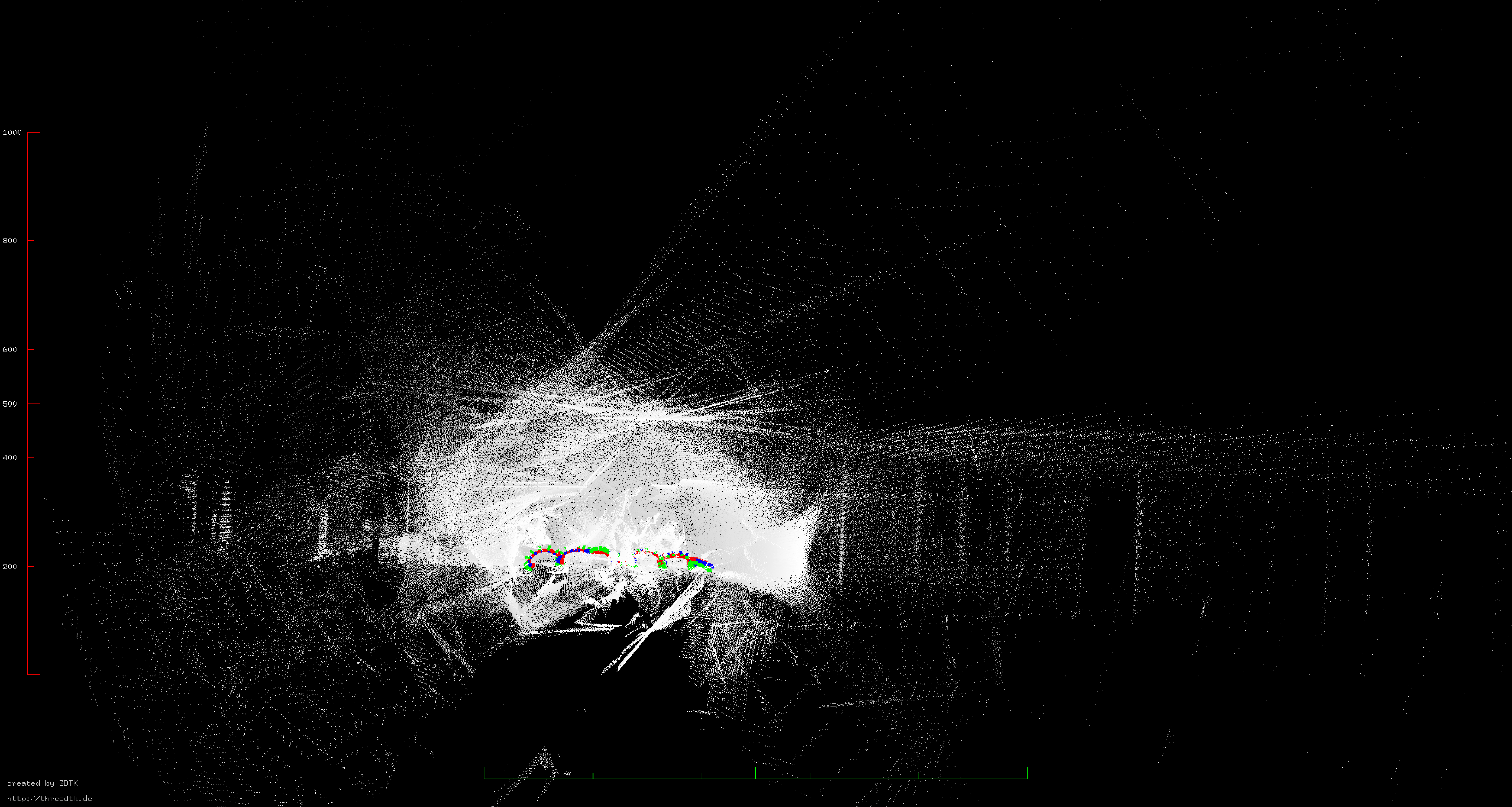}
      \end{minipage}%
    }%
    \hspace{2pt}%
    \subfloat[]{%
      \begin{minipage}[b]{0.485\linewidth}
        \centering
        \includegraphics[width=\linewidth]{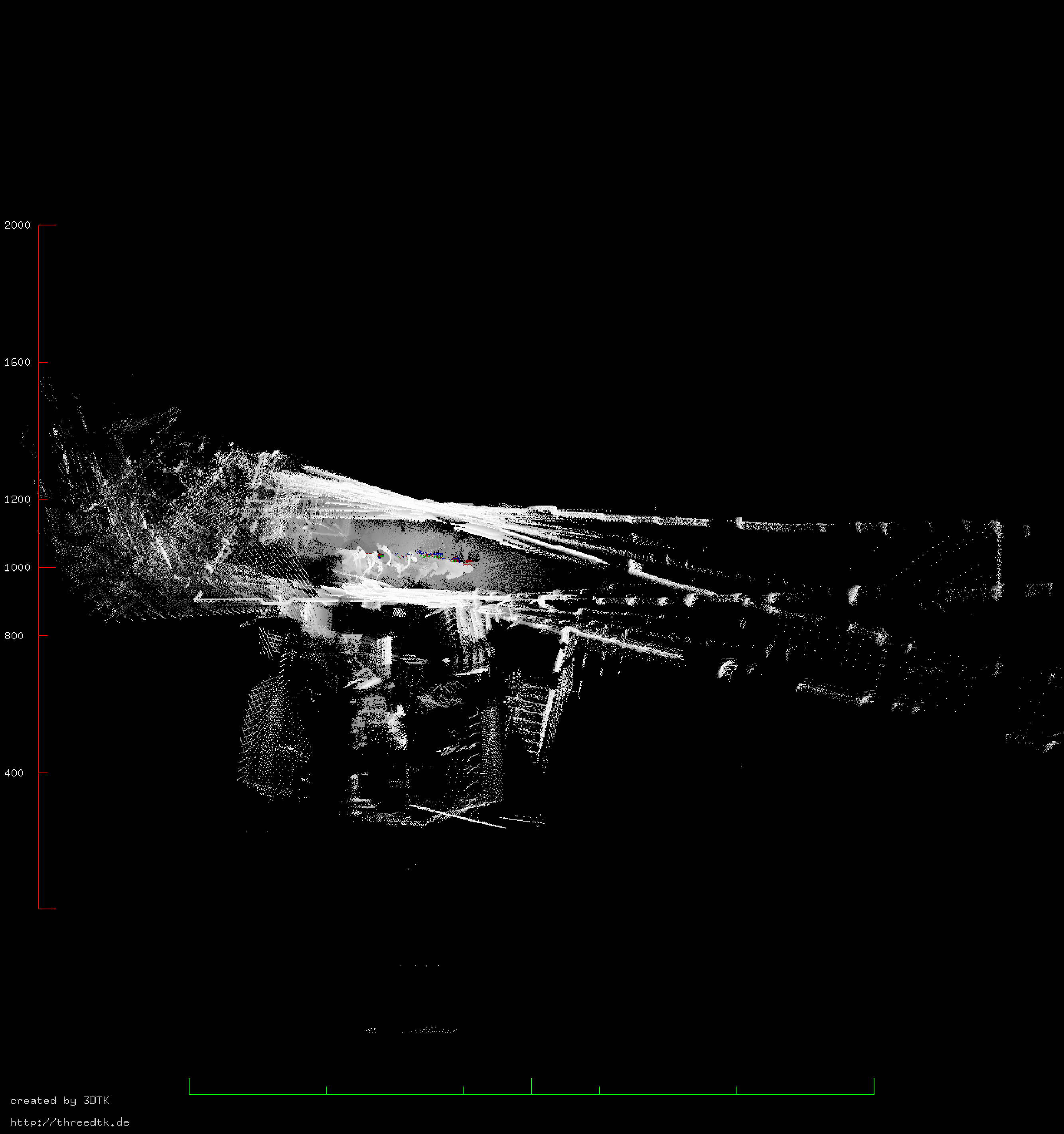}\vspace{1pt}
        \includegraphics[width=\linewidth]{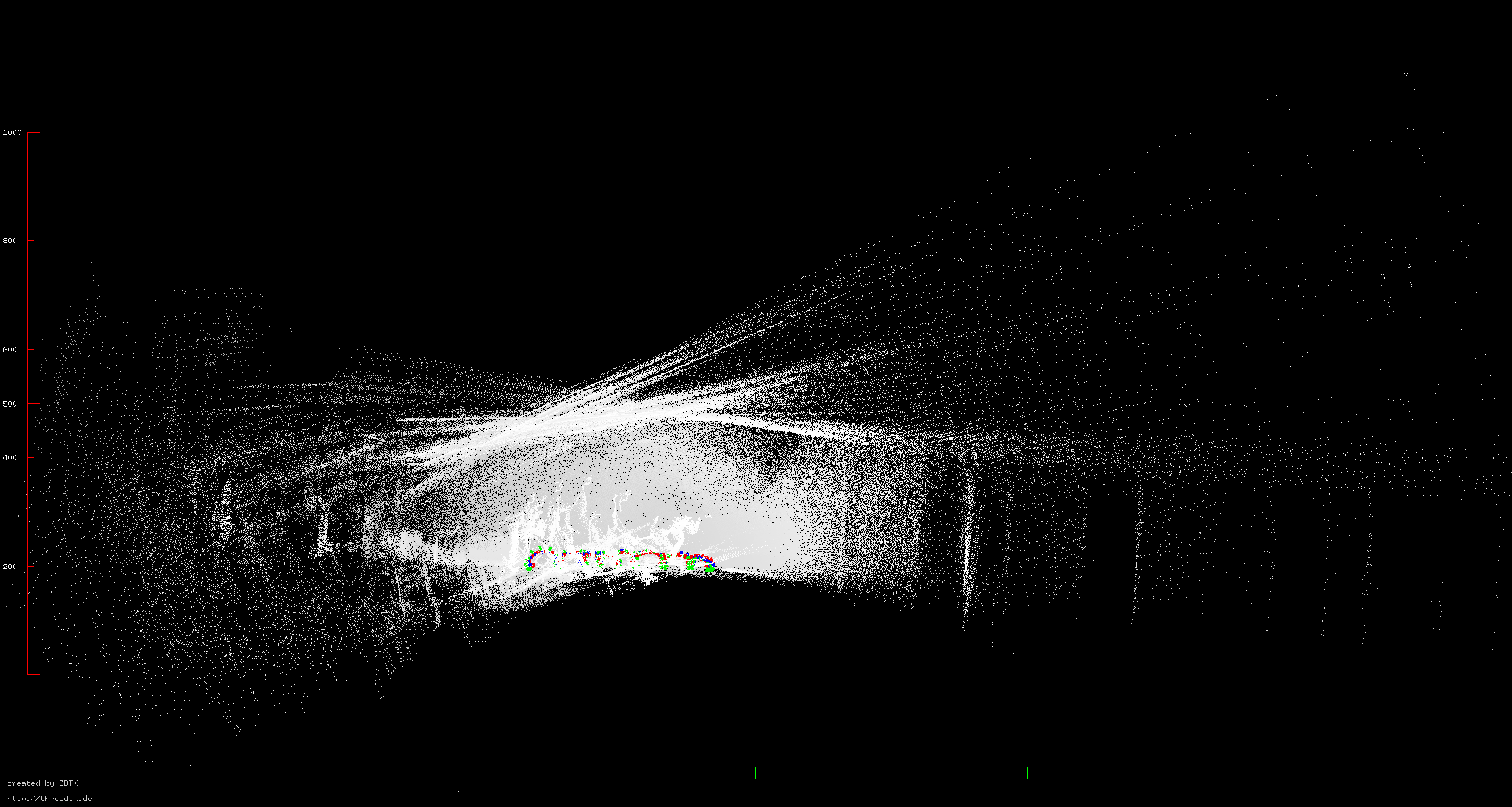}
      \end{minipage}%
    }%
  \end{minipage}
  \begin{minipage}[t]{0.485\textwidth}
    \centering
    \subfloat[]{%
      \begin{minipage}[b]{0.485\linewidth}
        \centering
        \includegraphics[width=\linewidth]{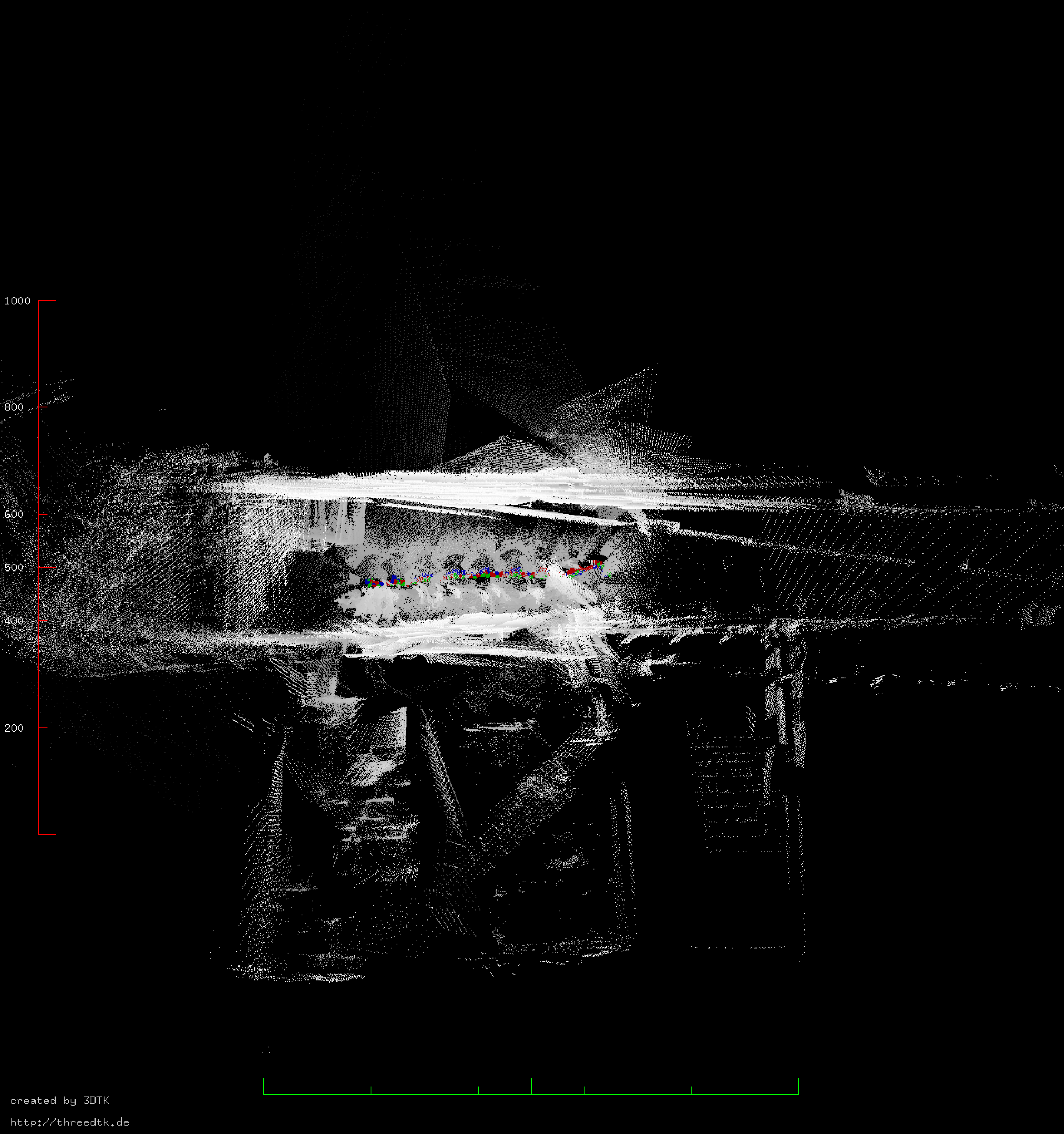}\vspace{1pt}
        \includegraphics[width=\linewidth]{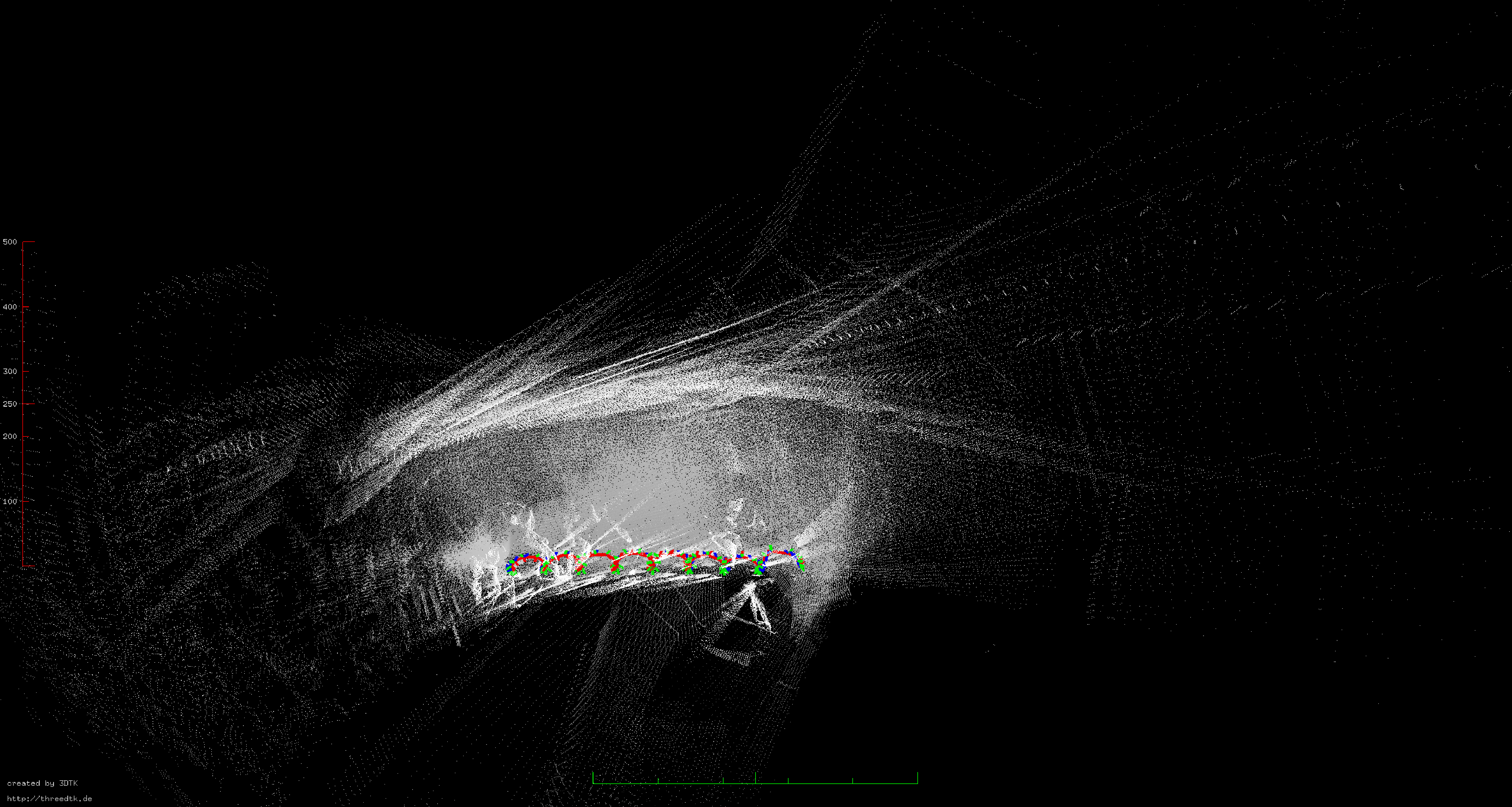}
      \end{minipage}%
    }%
    \hspace{2pt}%
    \subfloat[]{%
      \begin{minipage}[b]{0.485\linewidth}
        \centering
        \includegraphics[width=\linewidth]{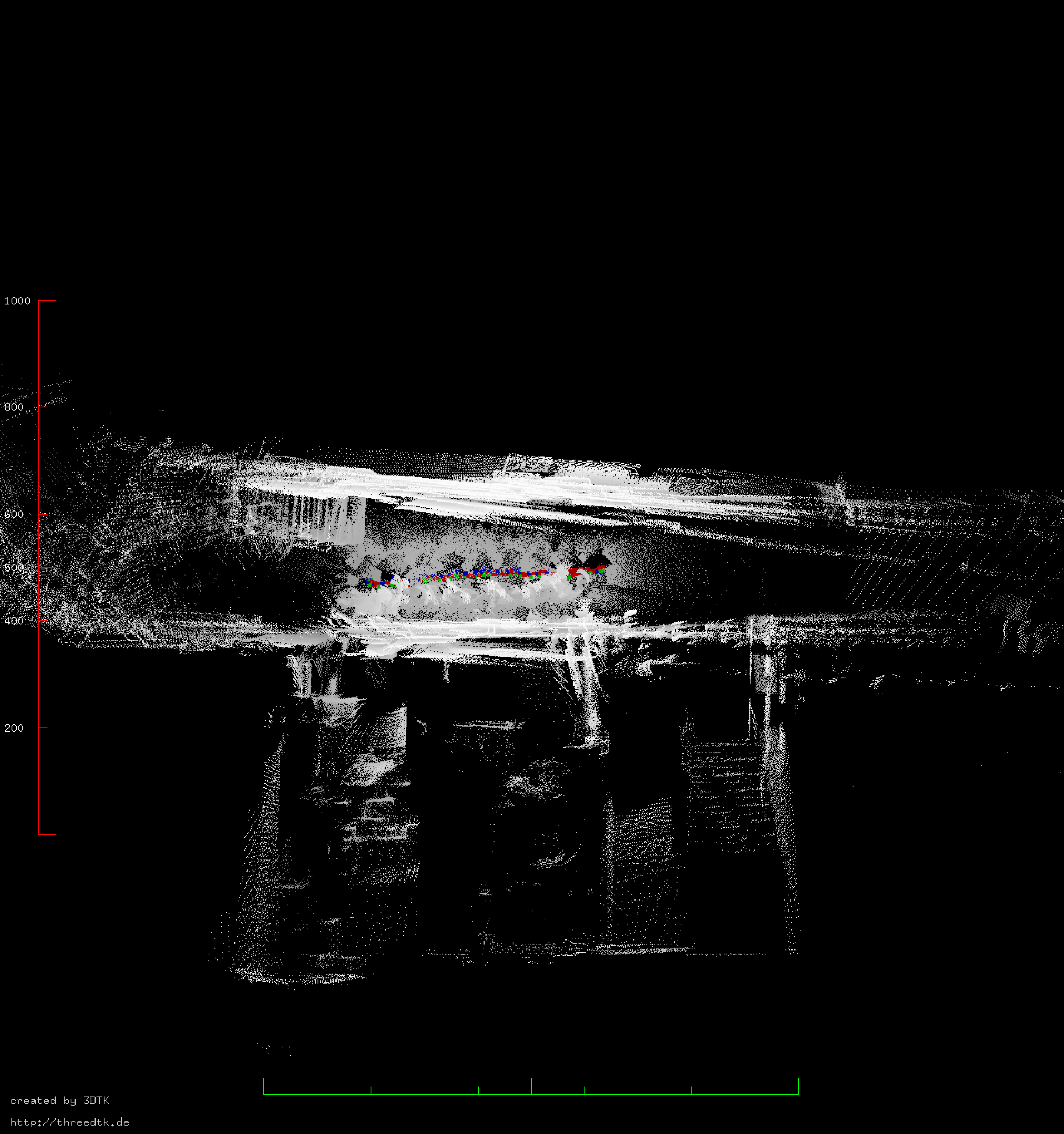}\vspace{1pt}
        \includegraphics[width=\linewidth]{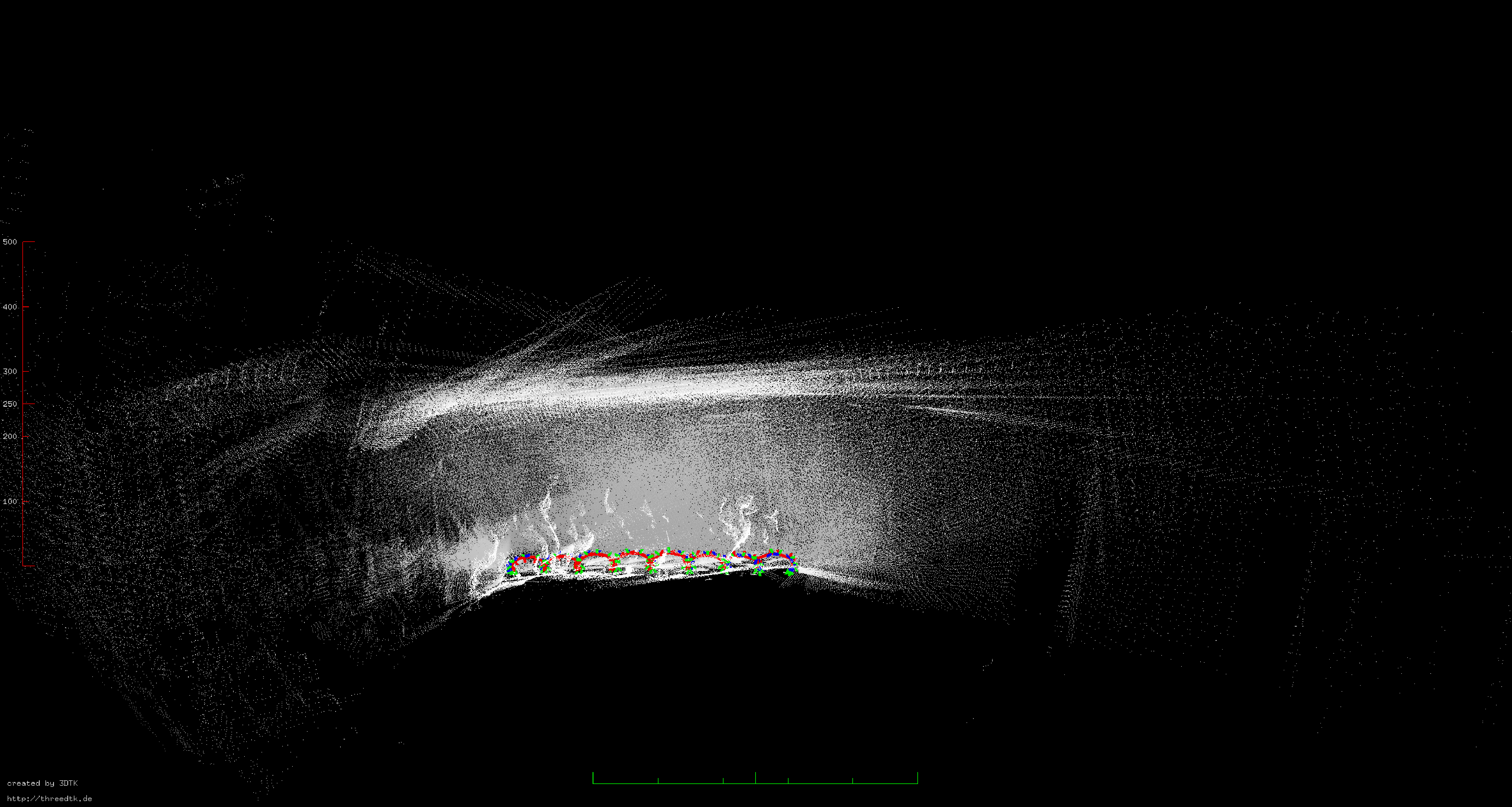}
      \end{minipage}%
    }%
  \end{minipage}

  \vspace{6pt}

  \begin{minipage}[t]{0.485\textwidth}
    \centering
    \subfloat[]{%
      \begin{minipage}[b]{0.485\linewidth}
        \centering
        \includegraphics[width=\linewidth]{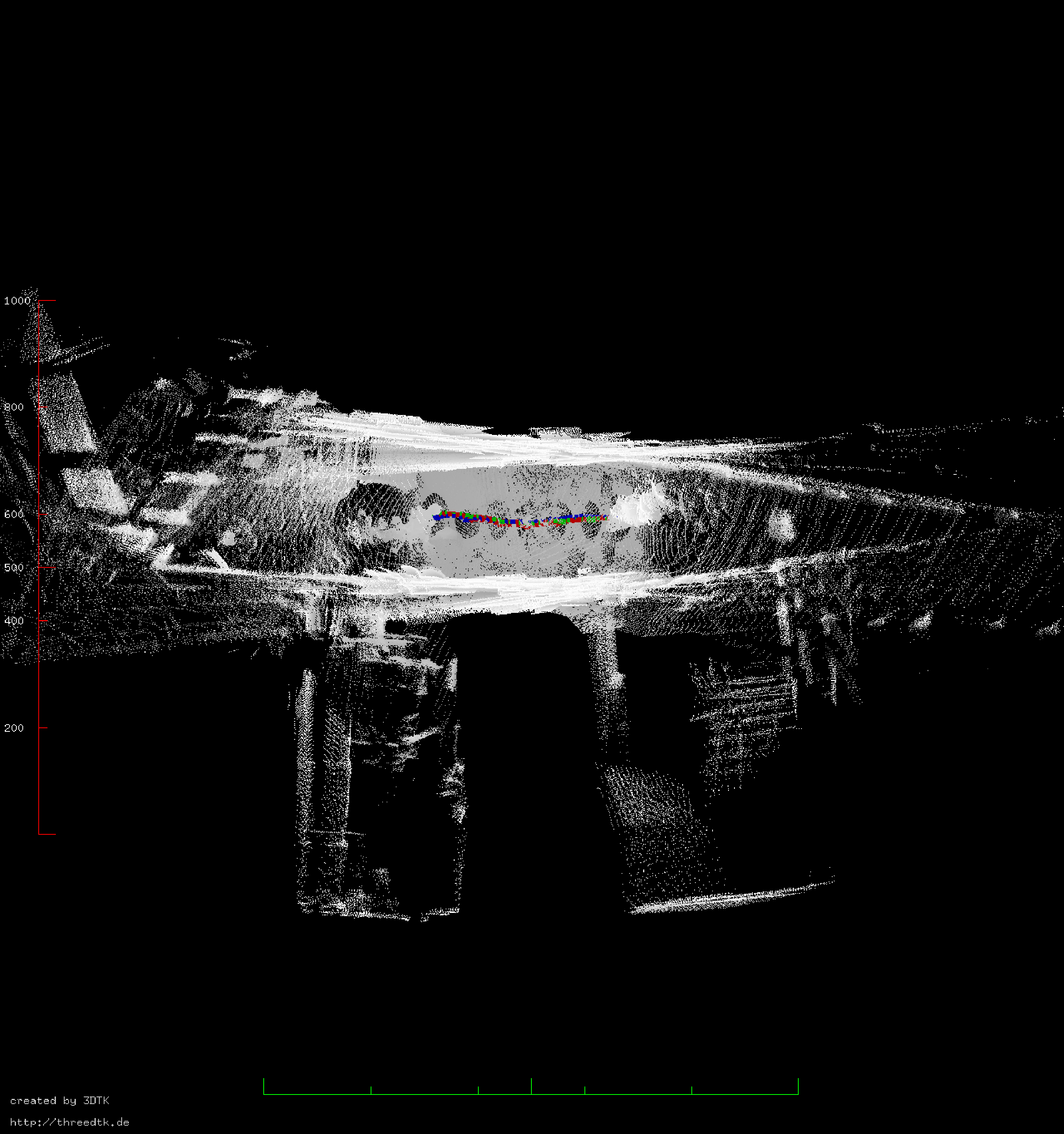}\vspace{1pt}
        \includegraphics[width=\linewidth]{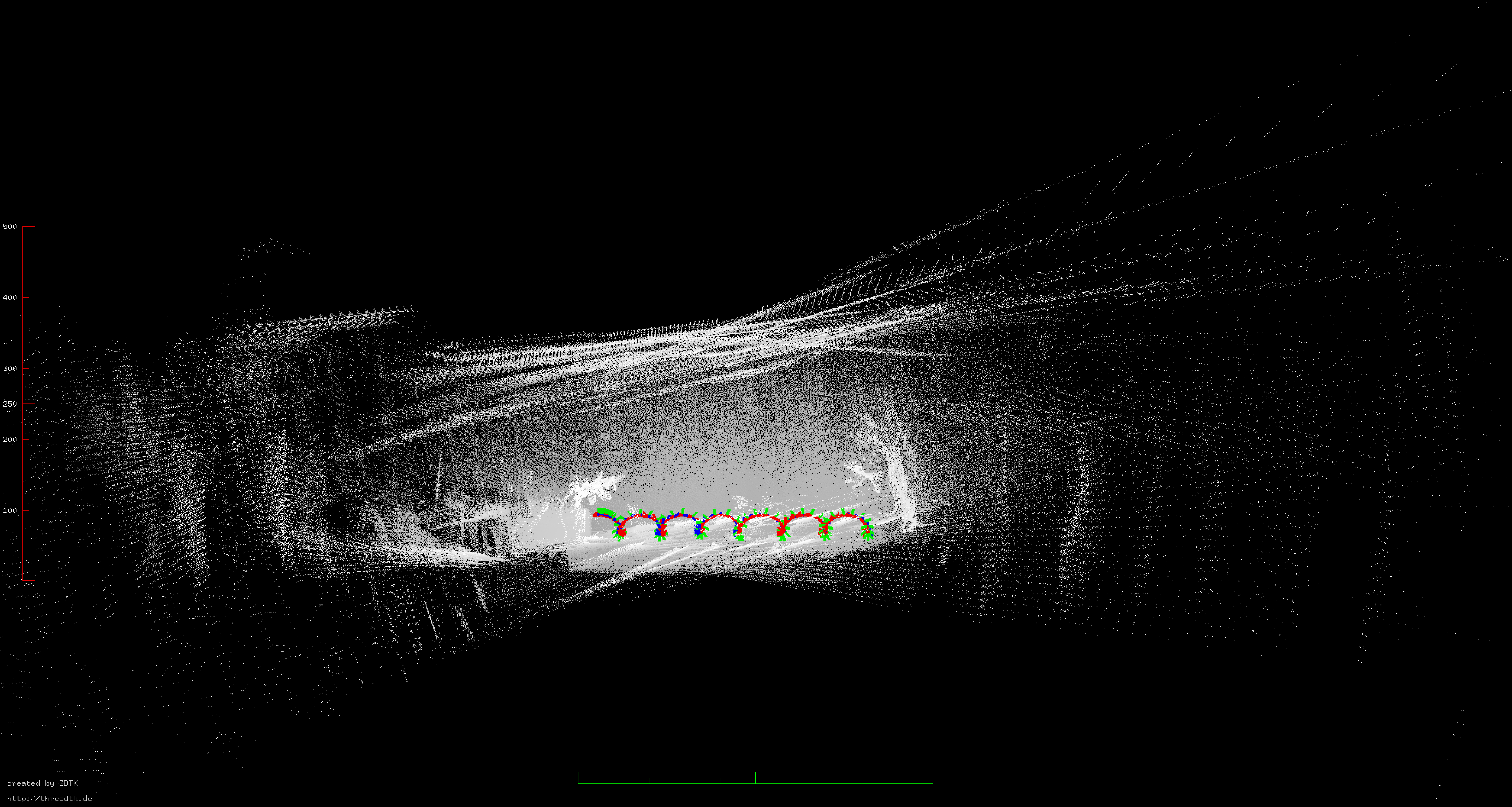}
      \end{minipage}%
    }%
    \hspace{2pt}%
    \subfloat[]{%
      \begin{minipage}[b]{0.485\linewidth}
        \centering
        \includegraphics[width=\linewidth]{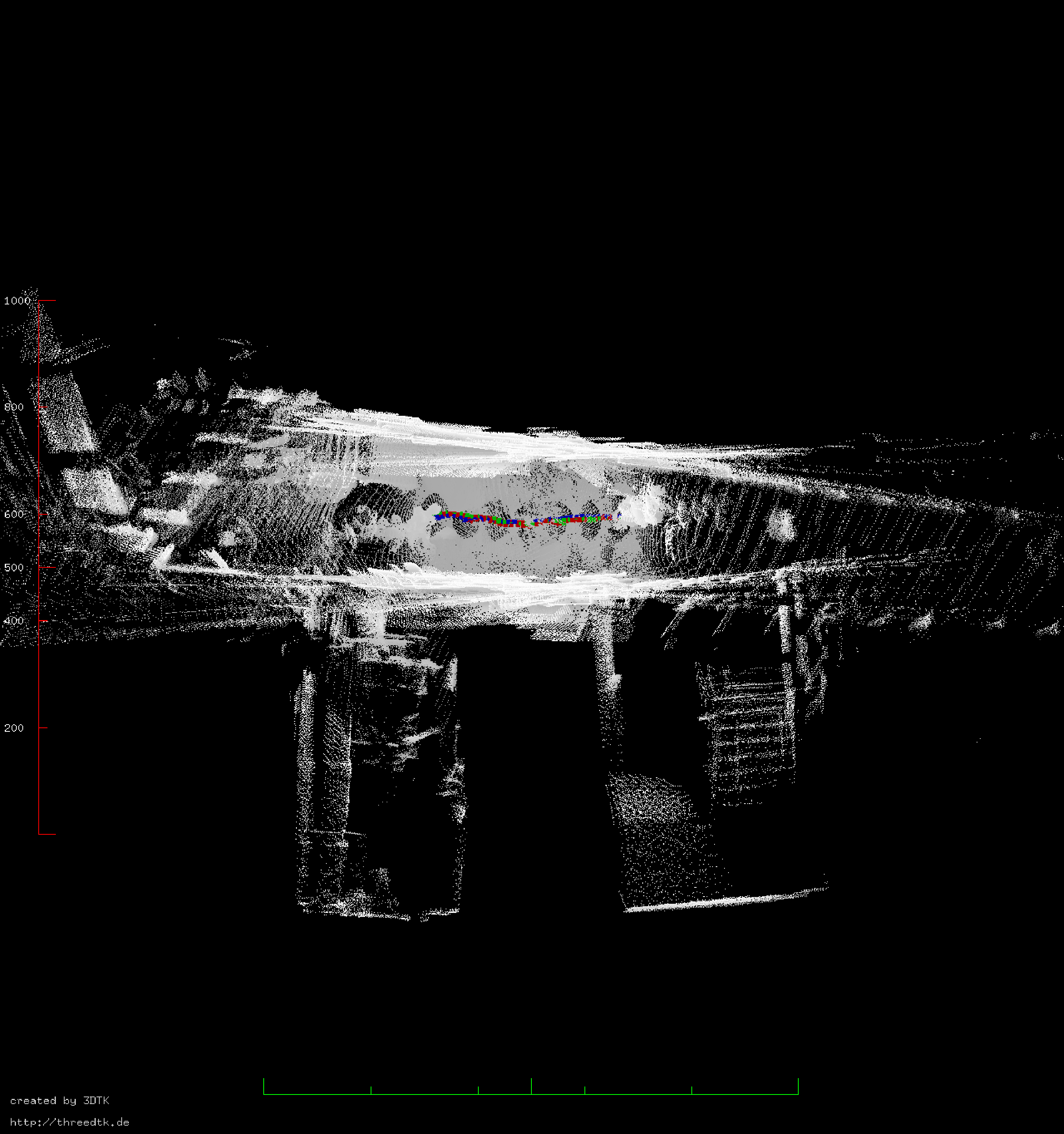}\vspace{1pt}
        \includegraphics[width=\linewidth]{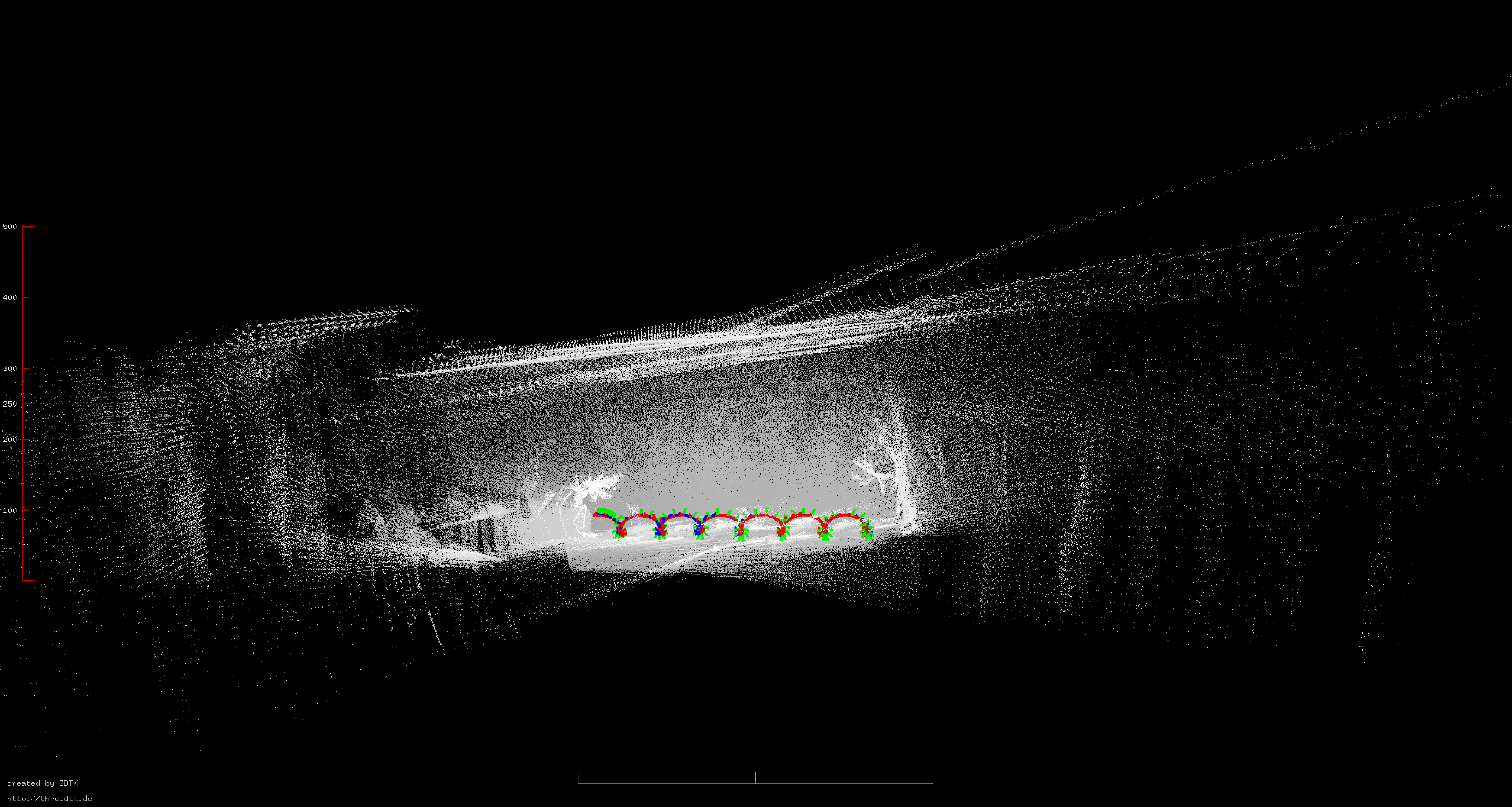}
      \end{minipage}%
    }%
  \end{minipage}
  \begin{minipage}[t]{0.485\textwidth}
    \centering
    \subfloat[]{%
      \begin{minipage}[b]{0.485\linewidth}
        \centering
        \includegraphics[width=\linewidth]{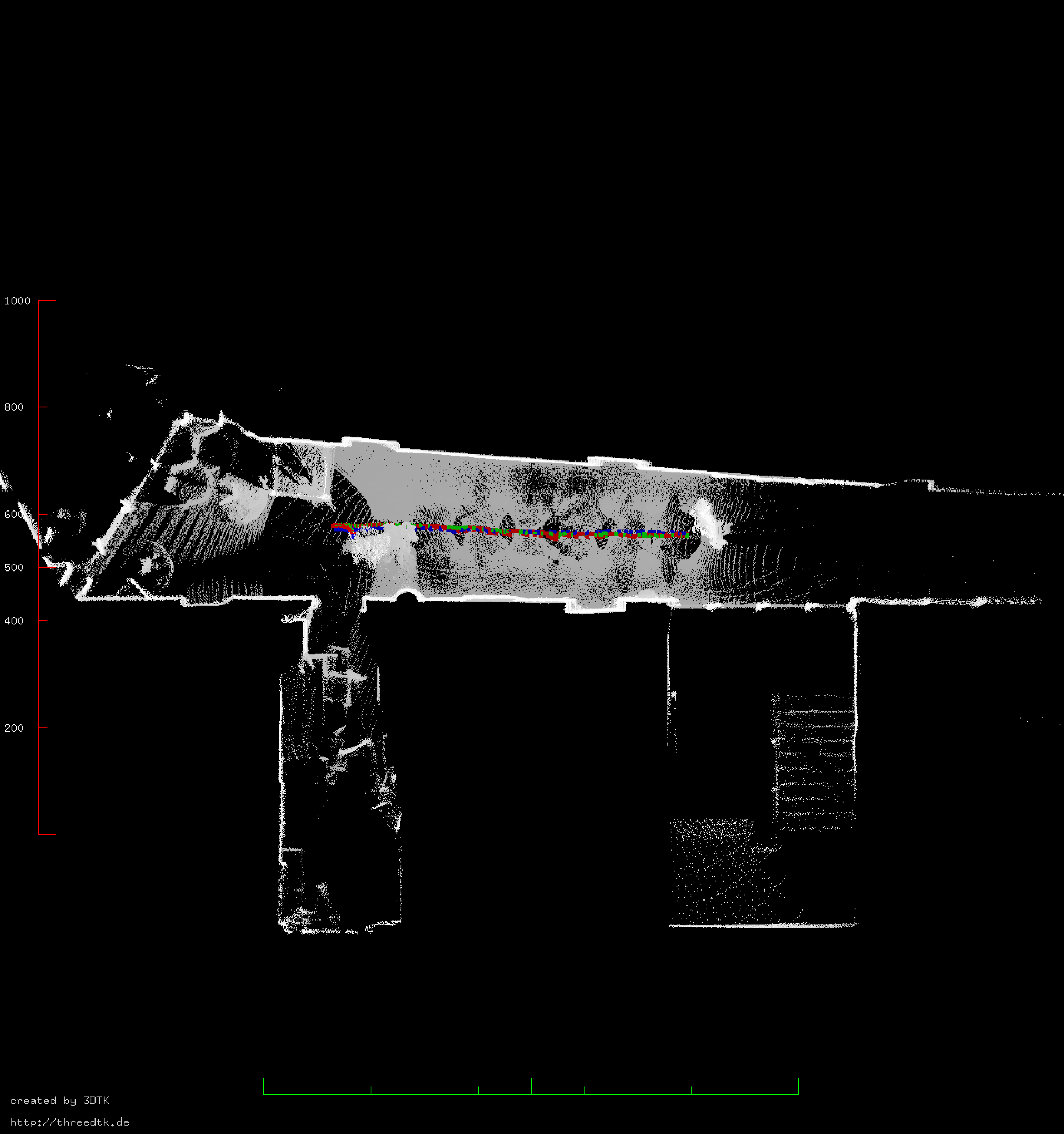}\vspace{1pt}
        \includegraphics[width=\linewidth]{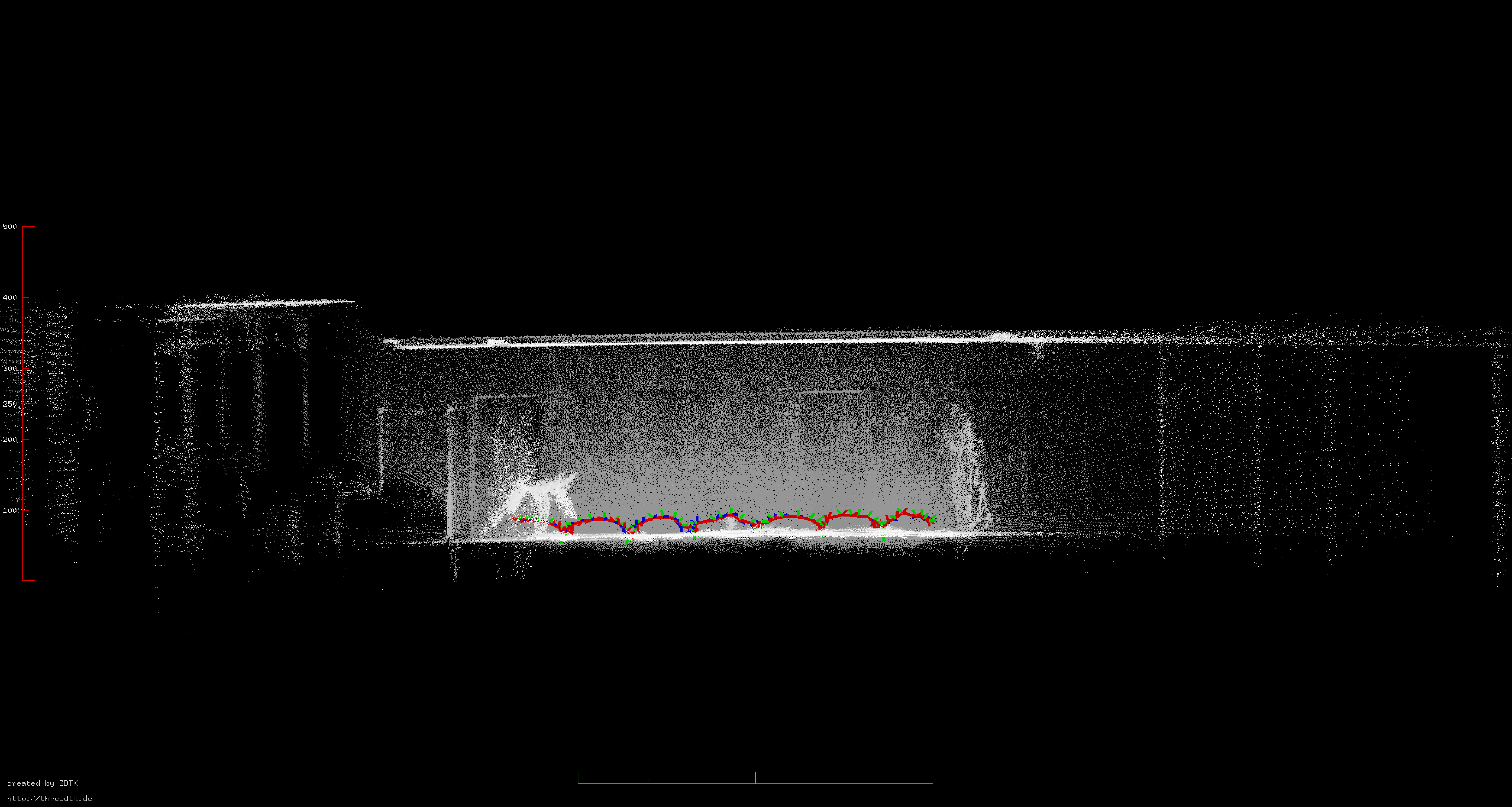}
      \end{minipage}%
    }%
  \end{minipage}
  \caption{Comparison of resulting point clouds from a spherical mobile mapping system when applying the attitude estimations of the ``Autogain''-filter with and without the proposed compensation.
  (Above rows:) Truncated orthographic birds-eye perspective with cut ceiling.
  (Below rows:) Truncated orthographic horizontal perspective with one cut wall.
  (a:) Single off-centered IMU without motion acceleration compensation.
  (b:) Single off-centered IMU with compensation. Same point data as in (a).
  (c:) Single centered IMU without compensation.
  (d:) Single centered IMU with compensation. Same point data as (c).
  (e:) Multiple, symmetrically placed IMU's without compensation.
  (f:) Multiple, symmetrically placed IMU's with compensation. Same point data as in (e).
  (g:) Reference point cloud available from scan matching.}\label{fig:pointcompare}
\end{figure*}
Figure~\ref{fig:pointcompare} shows the resulting point clouds when applying the attitude estimations with and without compensation.
The point clouds in the left columns (a) and (b) correspond to the case of a single non-centered IMU, the point clouds in the columns (c) and (d) correspond to the case of a single centered IMU, and the point clouds in the columns (e) and (f) correspond to the case of multiple symetrically placed IMUs.
In a similar fashion as before, (a), (c), and (e) show the point clouds without compensation, and (b), (d), and (f) show the point clouds with compensation.
The most right column (g) shows a reference point cloud of the environment.
The largest improvement is visible in the single IMU cases, where the point clouds in (a) and (c) are significantly distorted and the point clouds in (b) and (d) are more structured and resembles the true geometry of the environment.
\begin{figure}
  \centering
  \includegraphics[width=\linewidth]{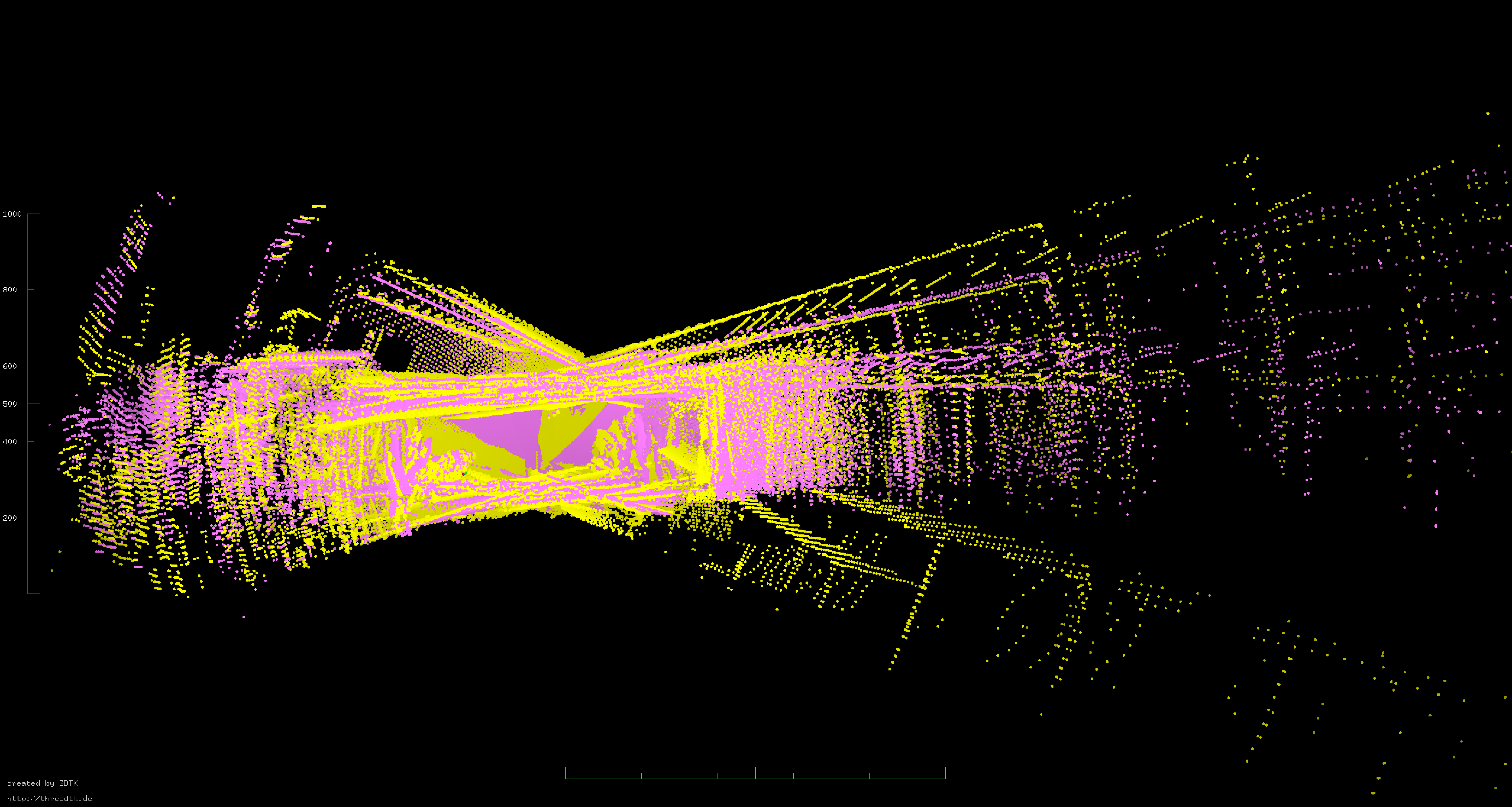}
  \caption{Detailed comparison of the resulting point clouds with and without the proposed compensation for symetrically placed IMUs.
  The view is a truncated orthographic horizontal perspective with one cut wall. 
  (Yellow:) Without compensation, corresponding to Figure~\ref{fig:pointcompare}~(e). 
  (Magenta:) With compensation, corresponding to Figure~\ref{fig:pointcompare}~(f).}\label{fig:pointzoom}
\end{figure}
The improvement in the multiple IMU case is less visible, but still noticeable when looking at the details of the point cloud in Figure~\ref{fig:pointzoom}:
The magenta colored point cloud, corresponding to active compensation, shows overall less angular error than the yellow colored one.  

\section{Conclusions}
In this work we present a method for calibration and compensation of IMUs that are mounted away from the center of rotation and therefore measure centripetal- and tangential-accelerations in addition to gravity. 
The method combines intrinsic calibration of the accelerometer and gyroscope triads with an estimation of the displacement vector to the center of rotation and, in the multi-IMU case, the relative orientations between sensors. 
Using these parameters, we correct the accelerometer measurements online by subtracting the motion-induced acceleration terms predicted from the gyroscope data.
The evaluation indicates that this correction improves the gravity measurements used by downstream attitude estimators. 
In a Monte-Carlo simulation, the calibration successfully recoveres the sensor transformations. 
On synthetic trajectories, the compensated measurements drastically reduces the attitude estimation error for state-of-the-art attitude filters. 
On the real-world spherical mobile mapping system, the improvement is most pronounced for a single off-centered IMU, where the compensated measurements leads to more accurate attitude estimates and thus, visibly less distorted point clouds. 
It is also possible to compensate the effect of motion-induced acceleration by symmetrically placing multiple IMUs around the center of rotation, and averaging their measurements.
Even in this case, when applying the proposed compensation method, the improvement is small but still observable.
Needless to say, a lot of work remains to be done.
The real-world experiments suggest that unmodelled sensor non-linearities exist. 
Furthermore, the extended model for spherical systems, which considers translational effects due to rolling on the ground, additionally requires an estimate of the surface normal in the local sensor frame.
We also want to evaluate the method on a broader range of systems and trajectories.
In future work we also plan implement tighter integration of the compensation term into the estimators themselves, e.g., in a Kalman filter. 
Nevertheless, the proposed method enables more accurate attitude estimation for systems with off-centered IMUs, which are common in countless applications such as mobile mapping, robotics, or in-orbit spacecraft.

\bibliographystyle{IEEEtran_doi}
\bibliography{root}

@techreport{deadalus,
  author    = {Angelo Pio Rossi and Francesco Maurelli and Vikram Unnithan and Hendrik Dreger and Kedus Mathewos and Nayan Pradhan and Dan-Andrei Corbeanu and Riccardo Pozzobon and Matteo Massironi and Sabrina Ferrari and Claudia Pernechele and Lorenzo Paoletti and Emanuele Simioni and Pajola Maurizio and Tommaso Santagata and Dorit Borrmann and Andreas N{\"u}chter and Anton Bredenbeck and Jasper Zevering and Fabian Arzberger and Camilo Andr{\´e}s Reyes Mantilla},
  title     = {{DAEDALUS - Descent And Exploration in Deep Autonomy of Lava Underground Structures}},
  institution = {Institut f{\"u}r Informatik},
  series    = {Forschungsberichte in der Robotik = Research Notes in Robotics},
  number    = {21},
  pages     = {188},
  doi        = {10.25972/OPUS-22791},
  year      = {2021},
}

@INPROCEEDINGS{euston2008complementary,
  author={Euston, Mark and Coote, Paul and Mahony, Robert and Kim, Jonghyuk and Hamel, Tarek},
  booktitle={2008 IEEE/RSJ International Conference on Intelligent Robots and Systems}, 
  title={A complementary filter for attitude estimation of a fixed-wing UAV}, 
  year={2008},
  volume={},
  number={},
  pages={340-345},
  doi={10.1109/IROS.2008.4650766}}

@techreport{hung1979size,
  title={Size effect on navigation using a strapdown IMU},
  author={Hung, J. C. and Hunter, J. S. and Stripling, W. W. and White, H. V.},
  year={1979},
  type={Technical Report},
  institution={Army Missile Command, Technology Laboratory},
  address={Redstone Arsenal, AL},
  series={MIRADCOM},
  number={DRSMI-T-79-73},
  doi={10.21236/ADA075434}
}

@MISC{ibm1969vstol_alt,
  title = "{V/STOL Inertial Navigation with Radar Update Capability}",
  howpublished = {NASA Contractor Report, NAS12-610},
  type={Technical Report},
  year = 1969,
  month = aug,
  note = {\href{https://ntrs.nasa.gov/api/citations/19690027876/downloads/19690027876.pdf}{IBM Contract No. 69-NC7-027}}
}

@INPROCEEDINGS{1257247,
  author={Kraft, E.},
  booktitle={Sixth International Conference of Information Fusion, 2003. Proceedings of the}, 
  title={A quaternion-based unscented Kalman filter for orientation tracking}, 
  year={2003},
  volume={1},
  number={},
  pages={47-54},
  doi={10.1109/ICIF.2003.177425}}

@INPROCEEDINGS{julier1997new,
       author = {{Julier}, Simon J. and {Uhlmann}, Jeffrey K.},
        title = "{New extension of the Kalman filter to nonlinear systems}",
    booktitle = {Signal Processing, Sensor Fusion, and Target Recognition VI},
         year = 1997,
       editor = {{Kadar}, Ivan},
       series = {Society of Photo-Optical Instrumentation Engineers (SPIE) Conference Series},
       volume = {3068},
        month = jul,
        pages = {182-193},
          doi = {10.1117/12.280797},
}

@ARTICLE{10203029,
  author={Li, Peng and Zhang, Wen-An and Jin, Yuqiang and Hu, Zihan and Wang, Linqing},
  journal={IEEE Transactions on Instrumentation and Measurement}, 
  title={Attitude Estimation Using Iterative Indirect Kalman With Neural Network for Inertial Sensors}, 
  year={2023},
  volume={72},
  number={},
  pages={1-10},
  doi={10.1109/TIM.2023.3301066}}

@article{crassidis2007survey,
author = {Crassidis, John L. and Markley, F. Landis and Cheng, Yang},
title = {Survey of Nonlinear Attitude Estimation Methods},
journal = {Journal of Guidance, Control, and Dynamics},
volume = {30},
number = {1},
pages = {12-28},
year = {2007},
doi = {10.2514/1.22452}
}

@inbook{lefferts1982kalman,
author={Lefferts, Ern J and Markley, F Landis and Shuster, Malcolm D},
title = {Kalman filtering for spacecraft attitude estimation},
volume={5},
number={5},
journal={Journal of Guidance, control, and Dynamics},
doi = {10.2514/6.1982-70},
year={1982}
}

@INPROCEEDINGS{10160508,
  author={Chen, Kenny and Nemiroff, Ryan and Lopez, Brett T.},
  booktitle={2023 IEEE International Conference on Robotics and Automation (ICRA)}, 
  title={Direct LiDAR-Inertial Odometry: Lightweight LIO with Continuous-Time Motion Correction}, 
  year={2023},
  volume={},
  number={},
  pages={3983-3989},
  doi={10.1109/ICRA48891.2023.10160508}}

@ARTICLE{chen2022direct,
  author={Chen, Kenny and Lopez, Brett T. and Agha-mohammadi, Ali-akbar and Mehta, Ankur},
  journal={IEEE Robotics and Automation Letters}, 
  title={Direct LiDAR Odometry: Fast Localization With Dense Point Clouds}, 
  year={2022},
  volume={7},
  number={2},
  pages={2000-2007},
  doi={10.1109/LRA.2022.3142739}}

@article{chen2023direct,
  title={Direct lidar-inertial odometry and mapping: Perceptive and connective slam},
  author={Chen, Kenny and Nemiroff, Ryan and Lopez, Brett T},
  journal={arXiv preprint arXiv:2305.01843},
  year={2023},
  doi = {10.48550/arXiv.2305.01843}
}

@article{kalmanoriginal,
    author = {Kalman, R. E.},
    title = {A New Approach to Linear Filtering and Prediction Problems},
    journal = {Journal of Basic Engineering},
    volume = {82},
    number = {1},
    pages = {35-45},
    year = {1960},
    month = {03},
    issn = {0021-9223},
    doi = {10.1115/1.3662552}
}

@ARTICLE{6172226,
  author={Lee, Jung Keun and Park, Edward J. and Robinovitch, Stephen N.},
  journal={IEEE Transactions on Instrumentation and Measurement}, 
  title={Estimation of Attitude and External Acceleration Using Inertial Sensor Measurement During Various Dynamic Conditions}, 
  year={2012},
  volume={61},
  number={8},
  pages={2262-2273},
  doi={10.1109/TIM.2012.2187245}
}

@INPROCEEDINGS{6957468,
  author={Blachuta, Marian and Grygiel, Rafal and Czyba, Roman and Szafranski, Grzegorz},
  booktitle={2014 19th International Conference on Methods and Models in Automation and Robotics (MMAR)}, 
  title={Attitude and heading reference system based on 3D complementary filter}, 
  year={2014},
  volume={},
  number={},
  pages={851-856},
  doi={10.1109/MMAR.2014.6957468}}

@INPROCEEDINGS{hall2008quaternion,
  author={Hall, James K. and Knoebel, Nathan B. and McLain, Timothy W.},
  booktitle={2008 IEEE/ION Position, Location and Navigation Symposium}, 
  title={Quaternion attitude estimation for miniature air vehicles using a multiplicative extended Kalman filter}, 
  year={2008},
  volume={},
  number={},
  pages={1230-1237},
  doi={10.1109/PLANS.2008.4570043}}

@article{zhang2017attitude,
title = {Attitude measure system based on extended Kalman filter for multi-rotors},
journal = {Computers and Electronics in Agriculture},
volume = {134},
pages = {19-26},
year = {2017},
issn = {0168-1699},
doi = {10.1016/j.compag.2016.12.021},
author = {Tiemin Zhang and Yihua Liao}}

@INPROCEEDINGS{jing2017attitude,
  author={Jing, Xiaofei and Cui, Jiarui and He, Hongtai and Zhang, Bo and Ding, Dawei and Yang, Yue},
  booktitle={2017 29th Chinese Control And Decision Conference (CCDC)}, 
  title={Attitude estimation for UAV using extended Kalman filter}, 
  year={2017},
  volume={},
  number={},
  pages={3307-3312},
  doi={10.1109/CCDC.2017.7979077}}

@INPROCEEDINGS{9110127,
  author={Jiachong, Chang and Ya, Zhang and Zhuo, Wang and Chao, Liu and Yanyan, Wang},
  booktitle={2020 IEEE/ION Position, Location and Navigation Symposium (PLANS)}, 
  title={Optimal inner lever-arm parameters calibration method of high-precision FOG-IMU based on sinusoidal swing scheme}, 
  year={2020},
  volume={},
  number={},
  pages={734-739},
  doi={10.1109/PLANS46316.2020.9110127}
}

@ARTICLE{9423615,
  author={Zhang, Xin and Zhou, Changle and Chao, Fei and Lin, Chih-Min and Yang, Longzhi and Shang, Changjing and Shen, Qiang},
  journal={IEEE Transactions on Industrial Informatics}, 
  title={Low-Cost Inertial Measurement Unit Calibration With Nonlinear Scale Factors}, 
  year={2022},
  volume={18},
  number={2},
  pages={1028-1038},
  doi={10.1109/TII.2021.3077296}
}

@ARTICLE{guner2022novel,
  author={Guner, Ufuk and Dasdemir, Janset},
  journal={IEEE Sensors Journal}, 
  title={Novel Self-Calibration Method for IMU Using Distributed Inertial Sensors}, 
  year={2023},
  volume={23},
  number={2},
  pages={1527-1540},
  doi={10.1109/JSEN.2022.3227341}}

@article{chen2024robust,
  title={Robust attitude estimation for low-dynamic vehicles based on MEMS-IMU and external acceleration compensation},
  author={Chen, Jiaxuan and Cui, Bingbo and Wei, Xinhua and Zhu, Yongyun and Sun, Zeyu and Liu, Yufei},
  journal={Sensors},
  volume={24},
  number={14},
  pages={4623},
  year={2024},
  doi = {10.3390/s24144623},
  publisher={MDPI}
}

@article{wei2025robust,
title = {A robust adaptive error state Kalman filter for MEMS IMU attitude estimation under dynamic acceleration},
journal = {Measurement},
volume = {242},
pages = {116097},
year = {2025},
issn = {0263-2241},
doi = {10.1016/j.measurement.2024.116097},
author = {Xiaofeng Wei and Shiwei Fan and Ya Zhang and Wei Gao and Feng Shen and Xie Ming and Jian Yang},
}

@INPROCEEDINGS{imutk2014,
  author={Tedaldi, David and Pretto, Alberto and Menegatti, Emanuele},
  booktitle={2014 IEEE International Conference on Robotics and Automation (ICRA)}, 
  title={A robust and easy to implement method for IMU calibration without external equipments}, 
  year={2014},
  volume={},
  number={},
  pages={3042-3049},
  doi={10.1109/ICRA.2014.6907297},
}

@article{microlie,
  title={A micro lie theory for state estimation in robotics},
  author={Sola, Joan and Deray, Jeremie and Atchuthan, Dinesh},
  journal={arXiv preprint arXiv:1812.01537},
  doi={10.48550/arXiv.1812.01537},
  year={2018}
}

@article{algediff,
title = {Systematic comparison of numerical differentiators and an application to model-free control},
journal = {European Journal of Control},
volume = {62},
pages = {113-119},
year = {2021},
note = {2021 European Control Conference Special Issue},
issn = {0947-3580},
doi = {10.1016/j.ejcon.2021.06.020},
author = {Amine Othmane and Joachim Rudolph and Hugues Mounier}
}

@INPROCEEDINGS{trochoidal,
  author={Arzberger, Fabian and Nüchter, Andreas},
  booktitle={2024 IEEE/RSJ International Conference on Intelligent Robots and Systems (IROS)}, 
  title={On the 3D trochoidal motion model of LiDAR sensors placed off-centered inside spherical mobile mapping systems}, 
  year={2024},
  volume={},
  number={},
  pages={1070-1077},
  doi={10.1109/IROS58592.2024.10801675}
}

@INPROCEEDINGS{imujasper,
author={Zevering, Jasper and Bredenbeck, Anton and Arzberger, Fabian and Borrmann, Dorit and N{\"u}chter, Andreas},
booktitle={2021 IEEE International Conference on Multisensor Fusion and Integration for Intelligent Systems (MFI)}, 
title={IMU-based pose-estimation for spherical robots with limited resources}, 
year={2021},
volume={},
number={},
pages={1-8},
doi={10.1109/MFI52462.2021.9591183}
}

@Article{complementary,
AUTHOR = {Valenti, Roberto G. and Dryanovski, Ivan and Xiao, Jizhong},
TITLE = {Keeping a Good Attitude: A Quaternion-Based Orientation Filter for IMUs and MARGs},
JOURNAL = {Sensors},
VOLUME = {15},
YEAR = {2015},
NUMBER = {8},
PAGES = {19302--19330},
PubMedID = {26258778},
ISSN = {1424-8220},
DOI = {10.3390/s150819302},
}

@INPROCEEDINGS{madgwick,
  author={Madgwick, Sebastian O. H. and Harrison, Andrew J. L. and Vaidyanathan, Ravi},
  booktitle={2011 IEEE International Conference on Rehabilitation Robotics}, 
  title={Estimation of IMU and MARG orientation using a gradient descent algorithm}, 
  year={2011},
  volume={},
  number={},
  pages={1-7},
  doi={10.1109/ICORR.2011.5975346},
}

@ARTICLE{mahony,
  author={Mahony, Robert and Hamel, Tarek and Pflimlin, Jean-Michel},
  journal={IEEE Transactions on Automatic Control}, 
  title={Nonlinear Complementary Filters on the Special Orthogonal Group}, 
  year={2008},
  volume={53},
  number={5},
  pages={1203-1218},
  doi={10.1109/TAC.2008.923738},
}

@ARTICLE{qekf,
  author={Sabatini, A.M.},
  journal={IEEE Transactions on Biomedical Engineering}, 
  title={Quaternion-based extended Kalman filter for determining orientation by inertial and magnetic sensing}, 
  year={2006},
  volume={53},
  number={7},
  pages={1346-1356},
  doi={10.1109/TBME.2006.875664},
}

\section*{Appendix}

\subsection{Derivation of EMA parameter}\label{apx:ema}
The causal first-order exponential moving average (EMA) filter is:
\begin{align}
y[i] = (1 - \alpha) \cdot y[i-1] + \alpha \cdot x[i]\,, \quad \text{with}\quad \alpha \geq 0\,.
\end{align}
Taking the Z-transform and rearranging to find the transfer function:
\begin{align}
Y[z] = (1 - \alpha) \cdot z^{-1} Y[z] + \alpha \cdot X[z] \\
\Rightarrow H(z) = \frac{Y[z]}{X[z]} = \frac{\alpha}{1 - (1-\alpha)z^{-1}}
\end{align}
We obtain the frequency response by substituting $z = e^{j\omega}$ where $\omega = 2\pi f / f_s$:
\begin{align}
H(e^{j\omega}) = \frac{\alpha}{1 - (1-\alpha)e^{-j\omega}}
\end{align}
The magnitude response is:
\begin{align}
|H(e^{j\omega})| = \frac{\alpha}{|1 - (1-\alpha)e^{-j\omega}|}
\end{align}
Expanding the denominator using Eulers formula:
\begin{align}
|&1 - (1-\alpha)e^{-j\omega}| \\
&= \sqrt{[1 - (1-\alpha)\cos\omega]^2 + [(1-\alpha)\sin\omega]^2} \\
&= \sqrt{1 - 2(1-\alpha)\cos\omega + (1-\alpha)^2}
\end{align}
Therefore:
\begin{align}
|H(e^{j\omega})| = \frac{\alpha}{\sqrt{1 - 2(1-\alpha)\cos\omega + (1-\alpha)^2}}
\end{align}
At the cutoff frequency $\omega_c = 2\pi \frac{f_{\text{cut}}}{f_s}$, we want to set the half-power point where the power gain is $10\log_{10}(0.5) \approx -3$~dB.
Thus, we need to solve:
\begin{align}
|H(e^{j\omega_c})|^2 = 1/2\,,
\end{align}
which yields a quadratic polynomial in $\alpha$:
\begin{align}
-\frac{1}{2} &\alpha^2 + \left(\cos(\omega_c) - 1\right) \alpha + 1 - \cos(\omega_c) = 0\\
\Rightarrow\quad &\alpha = \sqrt{\cos(\omega_c)^2 - 4\cos(\omega_c) + 3} + \cos(\omega_c) - 1 \,.
\end{align}

\end{document}